\documentclass[10pt,twocolumn,letterpaper]{article}

\usepackage[pagenumbers]{cvpr} 

\usepackage{cuted} 

\usepackage{currfile} 
\usepackage{tabularx}
\usepackage{booktabs}
\usepackage{longtable}
\usepackage{multirow}
\usepackage{makecell}
\usepackage{xltabular}
\usepackage{float}
\usepackage[table]{xcolor}
\usepackage[most]{tcolorbox} 
\tcbset{promptbox/.style={enhanced, breakable, colback=black!3, colframe=black!65, coltitle=white, fonttitle=\bfseries\small, boxrule=0.5pt, arc=2pt, left=6pt, right=6pt, top=4pt, bottom=4pt}}
\usepackage{caption} 

\newcommand{\be}[1]{\textbf{#1}}

\definecolor{cvprblue}{rgb}{0.21,0.49,0.74}
\usepackage[pagebackref,breaklinks,colorlinks,allcolors=cvprblue]{hyperref}

\def\confName{CVPR}
\def\confYear{2027}

\title{PAIR: Bringing Household Robots into the Smart Home}

\author{
Junseong Pyo,
Hyeongjin Kim,
Jongwook Han,
Seungwon Kook,
Sungjib Lim,
Yohan Jo\textsuperscript{*}
\\
Graduate School of Data Science, Seoul National University
\\
{\tt\small \{junseongpyo, gudwls5789, johnhan00, cacu5615, saint7451, yohan.jo\}@snu.ac.kr}
\\
\textsuperscript{*}Corresponding author
}

\begin{document}
\maketitle 

\begin{strip}         
    \centering
    \includegraphics[width=\linewidth]{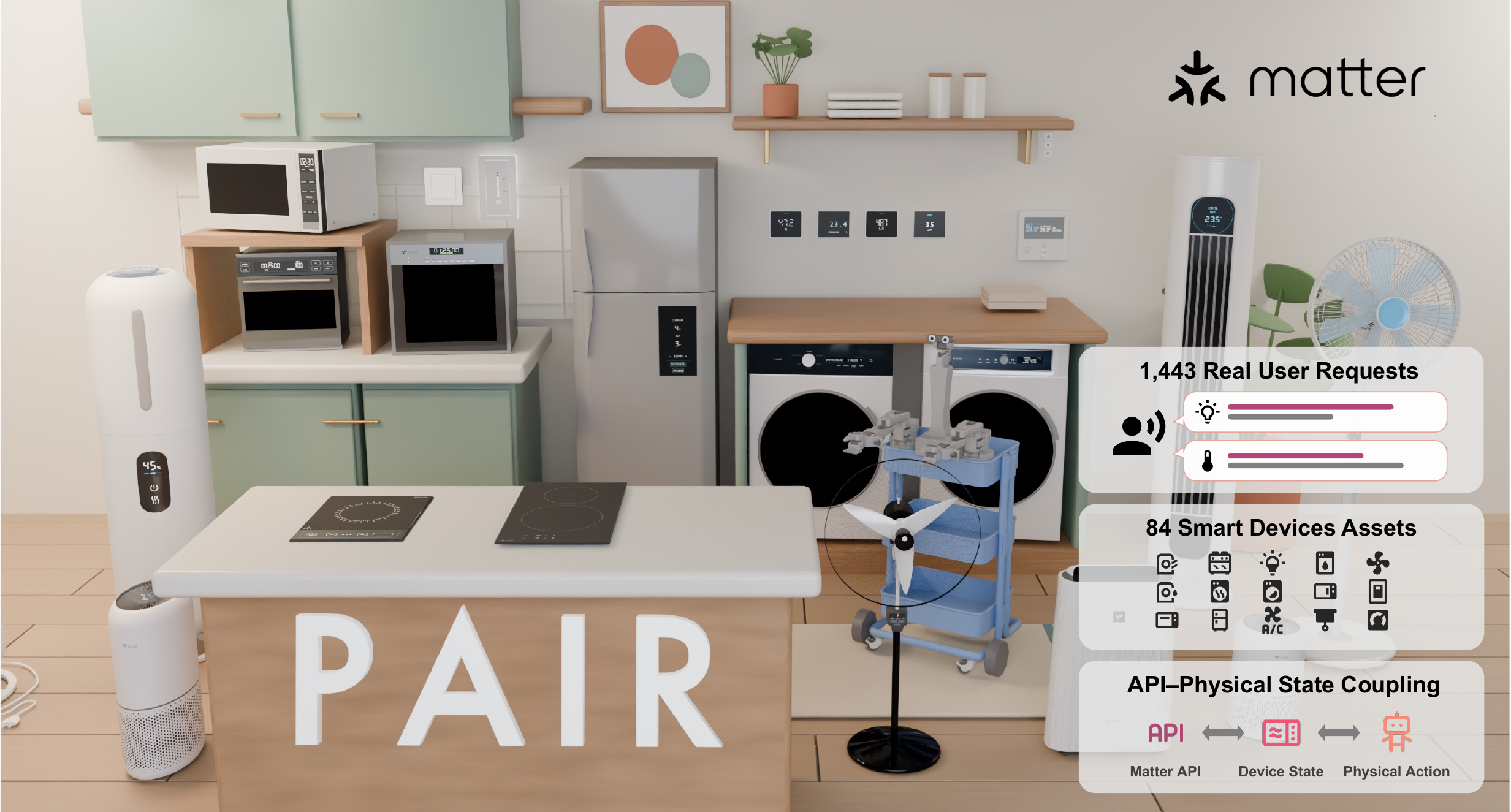}
    \captionsetup{hypcap=false}
    \captionof{figure}{\textbf{PAIR} enables household robots to interact with smart appliances through both physical controls and device APIs.}
    \captionsetup{hypcap=true}
    \label{fig:header}
\end{strip}

\begin{abstract}
  Household robots often need to operate appliances, and smart appliances offer both control panels and device APIs. Existing embodied AI simulation environments, however, model appliances solely as physical objects. We introduce PAIR, a smart-home simulation environment grounded in the Matter protocol~\footnote{https://csa-iot.org/all-solutions/matter/}, the industry standard for smart-home devices. PAIR provides 84 smart-device assets whose functional control panels and APIs act on a shared device state, with ambient conditions that evolve in response to both. From a survey of 1,443 requests, we construct a benchmark of 40 tasks with 340 human demonstrations for policy training. Whether run alone, with a high-level planner, or with an API-enabled planner, two fine-tuned VLA policies complete none of the 120 evaluation episodes. Robots rarely get far enough to use the API: even with oracle plans and API access, over 90\% of failed rollouts stall at navigating, grasping, or operating a door. With a perfect oracle in place of the robot's body, the best planner still gets through only 57.3\% of its plan. When choosing between pressing a button and calling an API, several planners go by whether they can read the panel, not by API latency.\footnote{We will release our code and data upon publication.}
\end{abstract}

\section{Introduction}
Building household robots that assist people with everyday tasks is a long-standing goal of embodied AI~\citep{challenges_robot_home, homerobot, behavior1k}. Performing these tasks often requires robots to operate household appliances by manipulating their controls, such as pressing buttons, turning knobs, and navigating touchscreens. Modern smart appliances, however, increasingly provide application programming interfaces (APIs), which allow external systems to read device states and send control commands through software~\citep{matter,alexa,google_home}.

Household tasks often involve both physical actions, such as navigation and object manipulation, and appliance operations that can be performed through APIs. Therefore, household robots need to flexibly select and combine physical actions and API calls to accomplish everyday tasks efficiently. For example, the request \textit{``Wash my shirt,''} requires a robot to physically load the shirt into the washing machine and close the door. Once the door is closed, the robot can start the cycle and monitor its progress through the API, without operating the control panel.

Studying how robots execute such tasks at scale calls for a simulation environment where device states change over time and where physical actions and API calls operate on the same devices. Existing simulation environments~\citep{libero, behavior1k, robocasa365} provide rich physics-based environments for robot manipulation, but they treat appliances as objects to be physically operated and do not consider smart-home settings, where devices report their states and accept commands through APIs. 

To bridge this gap, we introduce \textbf{PAIR} (Physical-API Integration Environment for Robots), a smart-home simulation environment that integrates physical and API-based device control. In PAIR, device states progress over time and respond to both physical actions and API calls, and running devices in turn change four ambient conditions of the home (temperature, illuminance, humidity, and air quality). We develop \textbf{84 smart-device assets}, each with a functional physical interface and an API, across 14 appliance categories and four sensor categories that report these ambient conditions. Device states, commands, and operational constraints are defined according to the Matter protocol~\citep{matter}, an industry standard for smart-device interoperability.

To ground evaluation in what people actually want from such a robot, we surveyed 129 participants about device-related household tasks they would delegate to a mobile manipulator capable of handling objects, operating appliances, and remotely controlling smart devices, and retained 1,443 requests from 108 participants after screening. We label each request with the skills it requires, using a taxonomy of robot skills in three categories, physical, perception, and human interaction (Table~\ref{tab:skills_taxonomy}). More than half of the requests (58.4\%) involve checking a device's status or waiting for an event, and APIs alone could complete only 39.6\% of them, while the remaining 60.4\% also require the robot's physical actions or perception. From these requests, we construct a benchmark of 40 tasks that cover the skills in our taxonomy and all device categories in PAIR. Every task requires object manipulation yet can be completed without API access, so the same task can be evaluated with and without APIs and across policy architectures.

Using this benchmark, we investigate how well current vision-language-action (VLA) policies handle such tasks. To adapt them to our robot and tasks, we collect 120 task-level demonstrations (three per task) and 220 demonstrations of atomic skills, all performed without API access. We then evaluate these policies alone, with a high-level planner, and with a planner that has API access, comparing task success and execution steps and analyzing how performance varies with the skills each task requires. None of the six VLA variants completes an evaluation task. API access raises device-operation success, yet with oracle plans and API access, more than 90\% of the failed rollouts still stop at a navigation, pick, or door subtask. Even with oracle physical execution, the best planner reaches only 57.3\% Progress, and several planners choose the API in response to panel visibility more consistently than to the stated API delay.

\begin{figure*}[t]
    \centering
    \includegraphics[width=\textwidth]{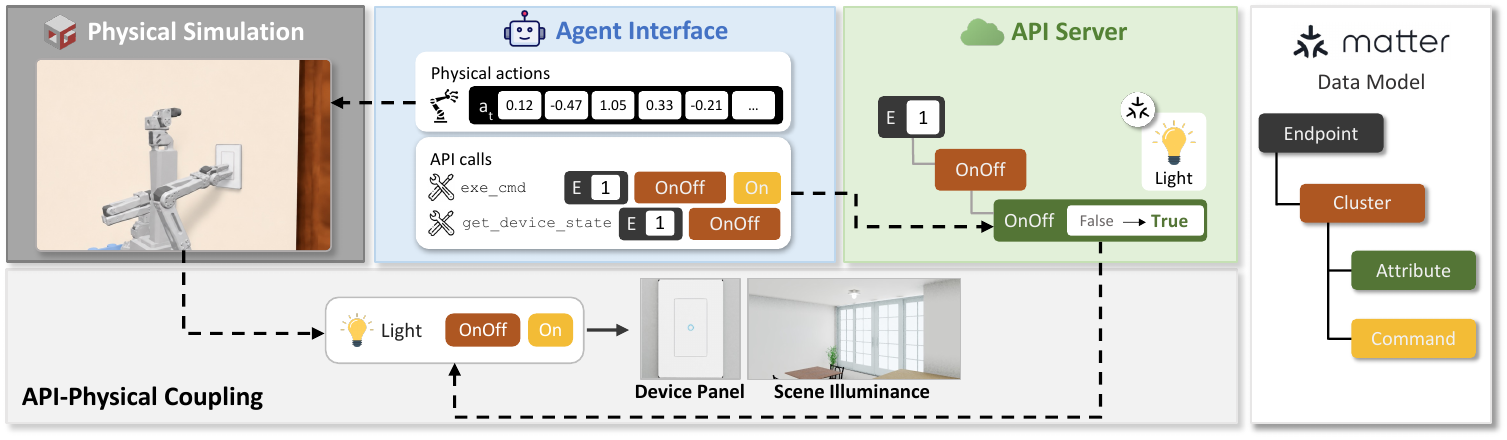}
    \caption{Architecture of PAIR. Agents interact with the physical simulation through physical actions and with a Matter-based API server through structured API calls. Both channels are coupled through a shared device state.}
    \label{fig:simulator}
\end{figure*}

\begin{figure}[!t]
    \centering
    {\setlength{\fboxsep}{0pt}\setlength{\fboxrule}{1pt}%
    \fcolorbox{black!25}{white}{\includegraphics[width=\dimexpr\columnwidth-2pt\relax]{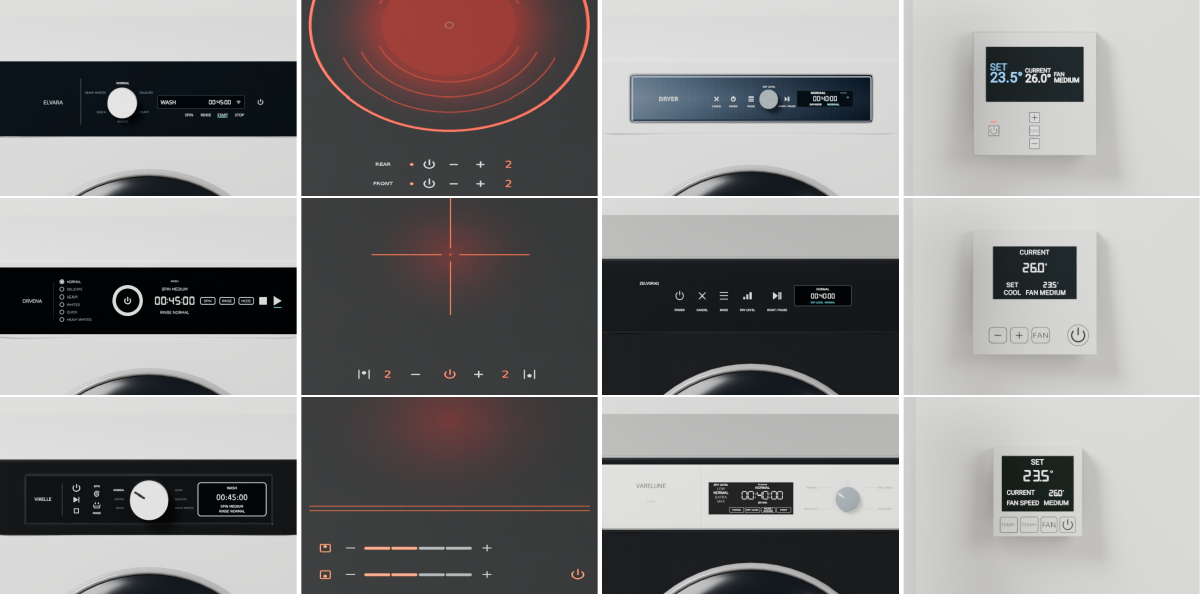}}}
    \caption{Representative control-panel variants in PAIR.}
    \label{fig:panel_variants}
\end{figure}

\section{Related Work}

\noindent\textbf{Household Simulation Environments for Robotics.}
Simulation environments have been a cornerstone of embodied AI research, enabling scalable training and evaluation in household environments while reducing the cost of real-world experiments. Early examples such as VirtualHome~\citep{virtualhome}, AI2-THOR~\citep{ai2_thor}, and ALFRED~\citep{alfred} established interactive 3D environments for household activities, but largely abstracted low-level object manipulation. Recent environments have expanded task diversity, dataset scale, and the realism and interactivity of articulated assets~\citep{behavior1k,robocasa,robocasa365}, enabling increasingly complex and physically grounded household tasks to be studied at scale. Despite these advances, existing environments treat appliances only as objects to be physically operated and do not consider smart-home settings, where devices report their states and accept commands through APIs.

\noindent\textbf{Benchmarks for Smart-Home Agents.}
Smart-home agent benchmarks evaluate household request understanding and connected-device control. Earlier benchmarks established LLM-based smart-home control by evaluating how agents translate natural-language household requests into sequences of API-based device operations~\citep{sage,sasha,homebench}. SimuHome~\citep{simuhome} advanced smart-home simulation by adopting the Matter protocol to model device states, commands, and operational constraints, while simulating their effects on ambient conditions over time. Moving beyond API-only control, OmniAct~\citep{omniact} routes between robot skills and IoT and API tools on real-world long-horizon tasks. Its robots, however, only pick, place, or navigate, its devices are controlled solely through simplified task-level tools, and it provides no simulation environment for training or evaluating robot policies. PAIR closes these gaps by building on SimuHome's Matter-based device model within a physics-based simulation environment, where the same device can be operated through its physical controls or its API.

\section{PAIR}
We present PAIR, a smart-home simulation environment that integrates physical and API-based device control, building on OmniGibson~\citep{behavior1k} for physical simulation. Figure~\ref{fig:simulator} illustrates the overall system architecture.

\subsection{Smart-Device Assets}
\label{sec:smart_device}
To support the realistic physical operation of smart-home devices, we construct 84 functional smart-device assets across 18 device categories, 14 of appliances and four of sensors. We source one to three appliance bodies per category, primarily from high-fidelity ArtVIP assets~\citep{artvip}, supplemented with appropriately licensed third-party assets. Inspired by real-world household appliances, we manually design and implement three functional control-panel variants (P1–P3) for each body, as exemplified in Figure~\ref{fig:panel_variants}. Each panel provides a control for every function and a display or indicator for every state of the underlying Matter-based device model. Compared with our measurements of commercial appliances, its buttons and knobs are somewhat larger and its displays are of similar size. We further evaluate the perceived realism of the resulting assets through human evaluation, obtaining an average realism score of 4.63/5 (Appendix~\ref{app:panel_realism}). Details on asset sources, licenses, panel measurements, and device-specific physical interfaces are provided in Appendix~\ref{app:smart_devices}.

\subsection{Agent Interface}
\label{sec:agent_interface}

Agents in PAIR interact with the environment through two channels: physical actions and API calls. For physical actions, agents receive visual observations of the environment and execute actions such as navigation, object manipulation, and device operation. Device indicators, panel displays, and sensor displays provide visual access to device states and ambient conditions.

For API calls, we use SimuHome's Matter-based device model~\citep{simuhome} to represent networked device states and commands. Agents can identify target devices, inspect their available attributes and commands, query device and sensor states, and issue device commands with the required arguments through structured APIs. Commands that cannot be executed given the device's current state and operational constraints are rejected without changing the device state. The full list of API tools is provided in Appendix~\ref{app:list_of_tools}.

\begin{table*}[t]
\centering
\small
\caption{Taxonomy of robot skills. The full taxonomy with definitions and examples is given in Table~\ref{tab:full_skills_taxonomy} (Appendix~\ref{app:skills_taxonomy}).}
\label{tab:skills_taxonomy}
\begin{tabularx}{\textwidth}{
    >{\raggedright\arraybackslash}p{60pt}
    >{\raggedright\arraybackslash}p{0.33\textwidth}
    >{\raggedright\arraybackslash}X
}
\toprule
\textbf{Category} & \textbf{Definition} & \textbf{Skills} \\
\midrule
\textbf{Physical} & Alters the physical world state. &
Pick and Place, Opening and Closing Doors, Twisting Knobs, Pressing Buttons, Device Operation, Navigation, etc. \\
\midrule
\textbf{Perception} & Updates the robot's knowledge of the world state without altering it. &
Status Check, Object Search and Identification, Object State Estimation, Event Detection, etc. \\
\midrule
\textbf{Human}\newline\textbf{Interaction} & Communicates with a person by receiving cues or conveying information. &
Spoken Cue Recognition, Informing \\
\bottomrule
\end{tabularx}
\end{table*}
 
\newsavebox{\surveytablebox}
\newsavebox{\surveypiebox}
\newlength{\piewidth}
\newlength{\surveyfiggap}
\begin{figure*}[t]
  \centering
  \setlength{\surveyfiggap}{2.5em}
  
    \newcommand{\surveytablerows}{%
        \toprule
        \multicolumn{2}{@{}l}{\textbf{Survey Overview}} \\
        \midrule
        Participants & 108 \\
        Requests & 1,443 \\
        \quad single-device & 1,169 \\
        \quad multi-device & 274 \\
        \midrule
        \multicolumn{2}{@{}l}{\textbf{Request Characteristics}} \\
        \midrule
        Involves status check or event detection & 58.4\% \\
        Not completable by APIs alone & 60.4\% \\
        \bottomrule
    }
  \begin{lrbox}{\surveytablebox}
    \small
    \begin{tabular}{@{}lr@{}}\surveytablerows\end{tabular}
  \end{lrbox}
  \setlength{\piewidth}{\dimexpr\textwidth-\wd\surveytablebox-\surveyfiggap\relax}
  \sbox{\surveypiebox}{\includegraphics[
    width=\piewidth,
    totalheight=\dimexpr\ht\surveytablebox+\dp\surveytablebox\relax,
    keepaspectratio
  ]{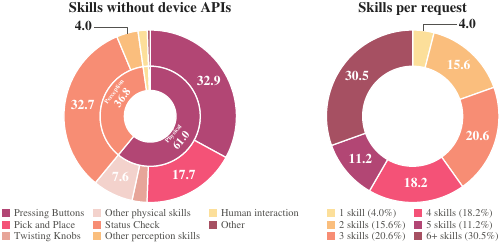}}
  \begin{lrbox}{\surveytablebox}
    \small
    \begin{tabular*}{\dimexpr\textwidth-\wd\surveypiebox-\surveyfiggap\relax}{@{\extracolsep{\fill}}lr@{}}
      \surveytablerows
    \end{tabular*}
  \end{lrbox}
  $\vcenter{\hbox{\usebox{\surveytablebox}}}$%
  \hspace{\surveyfiggap}%
  $\vcenter{\hbox{\usebox{\surveypiebox}}}$
  \caption{Survey statistics. Survey overview and request characteristics (left), distribution of required skills under physical-only control (middle), and number of skills per request (right). The middle chart counts each skill type at most once per request; the right chart counts each repeated physical action, such as pressing buttons for two settings, and each perception skill once.}
  \label{fig:survey_stats}
\end{figure*}

\subsection{API--Physical State Coupling}
\label{sec:state_coupling}

To maintain consistency between these two channels, PAIR uses a shared device state that can be updated through either channel. Changes to this state are reflected in the device's physical behavior and observable feedback, and, where applicable, drive the evolution of ambient conditions. We implement this coupling in both directions as follows.

\paragraph{Updates from API Calls.}
When an API command is executed, it updates the shared device state, with corresponding changes reflected in the device's physical behavior and observable feedback, such as control-panel displays and indicators. We additionally model network communication as a configurable one-way delay $d$, set to 200 ms by default based on message-delivery delays of up to 200 ms observed in prior measurements of Matter in an indoor home environment~\citep{matter_latency}. Accordingly, a command takes effect after $d$, and the response returns to the agent after a total round-trip delay of $2d$. A state query similarly reads the shared device state after $d$ and returns the observed value after $2d$.

\paragraph{Updates from Physical Operation.}
Physical operations update the shared device state locally according to the device's operational constraints, with the resulting changes reflected in the corresponding API-exposed state and observable feedback. Unlike API calls, physical operations are applied locally rather than transmitted over the network, so we introduce no additional communication delay for them. For example, pressing the power button on a smart light updates its power state, changes the corresponding observable feedback, and affects the room illuminance.

\section{Survey of Smart-Home Requests}
\label{sec:request_survey}

Benchmark tasks are often designed by researchers, which leaves open whether they reflect what people would actually ask a robot to do. BEHAVIOR-1K~\citep{behavior1k} addressed this by drawing activities from time-use surveys and asking people which ones they would want a robot to do. Yet, its activities assume a conventional home, not a smart home in which appliances can also be controlled remotely. We therefore surveyed 129 participants, asking them to imagine a robot capable of navigating the home, manipulating objects, operating appliances, and remotely controlling smart devices.

Each participant answered \textit{``What would you ask the robot to do?''} by writing single-device requests for devices assigned to them and multi-device requests involving two or more devices of their own choice, drawing on PAIR's 18 device categories (14 appliances and four sensors) and 124 household objects. For each request, they then marked the objects and device operations it involves, grounding the request in concrete objects and device functions. After excluding empty or uninterpretable responses and those that did not follow the instructions, we retained 1,443 requests from 108 participants (Appendix~\ref{app:survey_filtering}).

To characterize the skills these requests require, we construct a taxonomy of robot skills (Table~\ref{tab:skills_taxonomy}). We take physical skills from the foundational skills of RoboCasa~\citep{robocasa} and the definitions of the physical and perception categories from \citet{skill_survey}, and add skills that recur in the requests but appear in neither source, such as Spoken Cue Recognition. Because the physical skills needed to operate a device depend on its control panel, we label appliance operations with a generic Device Operation skill. We use GPT-6.1 Terra~\citep{gpt61terra} to label each request with the skills it requires, and validate the labels on 100 requests against the majority vote of three authors who annotated them independently (micro F1 = 0.881, $\kappa = 0.856$; Appendix~\ref{app:label_validation}).

We use the labels to characterize what the requests demand of a robot (Figure~\ref{fig:survey_stats}). To count the skills a request requires when every device is operated physically, we expand each Device Operation into the physical skills that the device's control panels in PAIR require (Section~\ref{sec:smart_device}; Appendix~\ref{app:skill_mapping}). Under this physical-only control, 96.0\% of the requests require multiple skills (95.3\% even among single-device requests), and 59.8\% require at least four (right). Perception accounts for 36.8\% of the required skills, most of it Status Check (32.7\%; middle). At the request level, 58.4\% of the requests involve checking a device's status or waiting for an event (left).

We then ask how much of this demand APIs can take over. If every device operation were instead available through an API, only 39.6\% of the requests could be completed through APIs alone, while the remaining 60.4\% would still require the robot's own physical actions or perception (left). People thus more often ask for tasks that combine the robot's physical capabilities with device control than for tasks that APIs alone can complete.

\section{Benchmark Design}
\label{sec:benchmark_design}

\begin{figure}[!t]
    \centering
    {\setlength{\fboxsep}{0pt}\setlength{\fboxrule}{1pt}%
    \fcolorbox{black!25}{white}{\includegraphics[width=\dimexpr\columnwidth-2pt\relax]{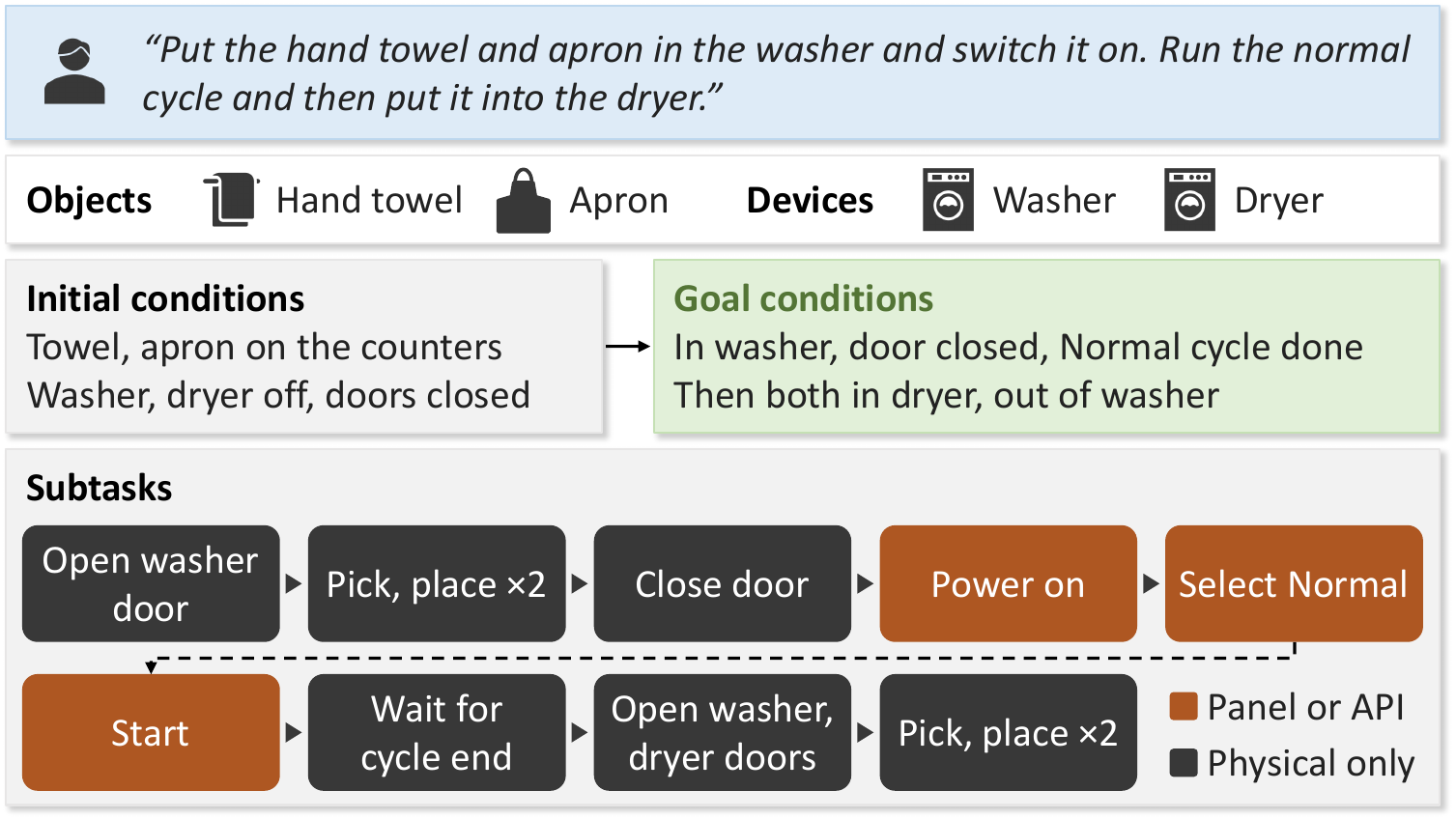}}}
    \caption{An example episode of a benchmark task.}
    \label{fig:task_example}
\end{figure}

\subsection{Task Design}
\label{sec:benchmark_task_design}
Each task in our benchmark specifies the required objects and devices, initial conditions, and goal conditions. Our benchmark focuses on how embodied agents combine physical actions with API access, and we therefore design all tasks around two principles: \textbf{physical necessity} and \textbf{physical-only solvability}. The first principle requires every task to include at least one step that no API can perform, such as loading dishes or closing a door, so that no task can be completed through APIs alone. The second principle requires every task to allow a complete physical-only solution, so that API access changes how a task is executed and not whether it can be solved. Based on these principles, we keep the 871 survey requests (Section~\ref{sec:request_survey}) that involve both a device operation and a step that no API can perform. After removing 327 that cannot be executed in the simulator and 148 by basic filtering, we select 40 tasks from the remaining 396 (Appendix~\ref{app:task_filtering}). Ten of the 40 tasks come from multi-device requests and involve two to four devices each, such as a washer and a dryer, or a sensor whose reading conditions an appliance operation.

\begin{table}[H]
  \caption{The five most frequent skills and devices in the 40 tasks.}
  \label{tab:represent_tasks_stats}
  \centering
  \small
  \setlength{\tabcolsep}{0pt}
  \begin{tabular*}{\columnwidth}{@{\extracolsep{\fill}} l r @{\hspace{10pt}} l r @{}}
    \toprule
    \textbf{Skills} & \textbf{Tasks} & \textbf{Devices} & \textbf{Tasks (\%)} \\
    \midrule
    Navigation            & 40 & Microwave oven & 7 (18\%) \\
    Pick and Place        & 39 & Dishwasher     & 6 (15\%) \\
    Device Operation      & 39 & Dryer          & 4 (10\%) \\
    Opening/Closing Doors & 23 & On/off light   & 4 (10\%) \\
    Status Check          & 15 & Oven           & 4 (10\%) \\
    \bottomrule
  \end{tabular*}
\end{table}

\begin{table*}
    \centering
    \small
    \caption{
        Main results on the 120 evaluation episodes. Subtask SR: percentage of oracle-plan subtasks completed, per skill. Progress: percentage of each episode's plan completed before its first uncompleted subtask. Best values in bold.
    }
    \label{tab:main_results}

    \setlength{\tabcolsep}{4pt}

    \begin{tabular*}{\textwidth}{
        @{\extracolsep{\fill}} l *{7}{c}
    }
        \toprule
        \multirow{3}{*}{\be{Method}} &
        \multicolumn{6}{c}{\be{Subtask SR (\%)} $\uparrow$} &
        \multirow{2}{*}{\be{Progress (\%)} $\uparrow$} \\
        \cmidrule(lr){2-7}
        &
        \be{Navigation} & \be{Pick} & \be{Place} &
        \be{Doors} & \be{Device Op.} & \be{All} & \\
        &
        \scriptsize(337) & \scriptsize(182) & \scriptsize(170) &
        \scriptsize(128) & \scriptsize(262) & \scriptsize(1,079) &
        \scriptsize(120 episodes) \\

        \midrule
        \rowcolor[HTML]{FFF3D9}
        \multicolumn{8}{l}{\textit{VLA Policies}} \\

        $\pi_{0.5}$
        & \textbf{38.0} & \textbf{12.6} & 0.0 & 6.2 & 0.0 & 14.7 & 8.9 \\
        \quad + HL
        & 35.6 & 11.5 & 0.0 & \textbf{7.0} & 0.4 & 14.0 & 7.8 \\
        \quad + HL (API)
        & 34.4 & 7.1 & 0.0 & 3.1 & 14.1 & \textbf{15.8} & \textbf{10.8} \\

        \addlinespace

        GR00T N1.7
        & 32.0 & 3.8 & \textbf{1.2} & 1.6 & 1.9 & 11.5 & 6.0 \\
        \quad + HL
        & 30.6 & 3.3 & 0.0 & 0.0 & 0.4 & 10.2 & 5.9 \\
        \quad + HL (API)
        & 30.0 & 5.5 & \textbf{1.2} & 1.6 & \textbf{14.5} & 14.2 & 10.3 \\
        \bottomrule
    \end{tabular*}
\end{table*}

\subsection{Episode Instantiation}
\label{sec:episode_instantiation}
For each task, we instantiate six episodes by randomizing the placement of task-relevant objects, the initial states of devices (e.g., open/closed, on/off), and the robot's initial base pose. Three of the six serve as training episodes for demonstration collection, and the remaining three are held out as evaluation episodes, yielding 120 episodes of each kind. An episode’s subtask sequence is generated by instantiating the task’s template according to its initial state (e.g., omitting the door-opening subtask when the door is initially open). Each subtask applies one skill from our taxonomy (Table~\ref{tab:skills_taxonomy}) to a single target, such as navigating to the dishwasher or picking up a plate. Figure~\ref{fig:task_example} shows one of the evaluation episodes. Table~\ref{tab:represent_tasks_stats} summarizes how many of the 40 tasks require each skill in their evaluation episodes, along with the most frequent devices. Details are provided in Appendix~\ref{app:episode_instantiation}.

\subsection{Demonstration Collection}
\label{sec:demonstration_collection}
To adapt VLA policies to PAIR, we collect 340 physical-only human demonstrations for fine-tuning. For each of the 120 training episodes, a human operator records one successful execution of the full task, giving three task-level demonstrations per task, each annotated with its subtask boundaries. We additionally collect 220 atomic demonstrations, each covering a single subtask in isolation, to give the policies more training data for individual physical skills. All demonstrations are collected on XLeRobot~\citep{xlerobot}, a low-cost, open-source dual-arm mobile manipulator that we use throughout this work. The demonstrations total 5.7 hours of robot execution. On average, a task-level demonstration lasts 130.5 s and consists of 9.9 subtasks (range 3--25) that span 4.7 of the five skills in Table~\ref{tab:represent_tasks_stats}. Details on teleoperation, subtask annotation, and data statistics are provided in Appendix~\ref{app:demonstration_collection}.

\section{Experiments}
\label{sec:experiments}

\subsection{Experimental Setup}

\noindent\textbf{Low-Level VLA.}
We fine-tune $\pi_{0.5}$~\citep{pi_05} and GR00T N1.7~\citep{gr00t} on the 340 PAIR demonstrations (Section~\ref{sec:demonstration_collection}). Each task-level demonstration is used both as a full trajectory with its task instruction and as subtask segments with their subtask instructions, and each atomic demonstration as a single subtask segment, so a policy can follow either kind of instruction, as in hierarchical VLA systems~\citep{pi_05,hirobot}. Training details are provided in Appendix~\ref{app:training_details}.

\noindent\textbf{High-Level Planner.}
We use Qwen3-VL-8B-Instruct as the high-level planner unless otherwise specified. The planner is queried every 5 s of simulated time, between which the VLA follows the current subtask. At each query, it receives the task instruction, the head and wrist camera images, the elapsed time, and a memory of its past subtasks. The high-level planner chooses one of five actions: (1) issue a new physical subtask to the VLA, (2) continue the current one, (3) wait, (4) end the episode, or (5) make an API call. API calling is available only in the + HL (API) setting (see Appendix~\ref{app:planner_prompt} for more details).

\noindent\textbf{Baselines.}
We evaluate each VLA in three variants. First, the standalone VLA executes the full task instruction without a high-level planner. In + HL, the planner instructs physical subtasks to the VLA without access to device APIs. In + HL (API), the planner is able to call device APIs.

\noindent\textbf{Evaluation Protocol.}
We run all variants \textbf{end to end} on the 120 evaluation episodes (three per task). An episode ends upon task success, robot fall, or exhaustion of an action budget set to 1.2 times the mean demonstration length for its task. Simulated time is paused during model inference so model latency does not affect the results.

To measure the effect of API access without upstream failures, we also evaluate \textbf{in isolation} each subtask that either channel can complete. Each subtask starts from the state in which a training demonstration begins it, restored by replaying the demonstration (100 subtasks, 78 device operations and 22 status checks). The planner receives the request and the demonstration's earlier subtasks, and each subtask must be completed within 90\,s of simulated time, either with physical subtasks only (+ HL) or with API access as well (+ HL (API)).

To diagnose planning and physical execution separately, we also replace one of the two with an \textbf{oracle} on one evaluation episode per task (40 episodes). In \textit{VLA execution with oracle plans}, the planner is replaced by the episode's ground-truth subtask sequence (Section~\ref{sec:episode_instantiation}). We evaluate this setting both with and without API access, with the oracle generating the correct API calls in the former. In \textit{high-level planning with oracle physical execution}, the VLA is replaced by scripted primitives that complete every physical subtask. Here we compare Qwen3-VL-4B/8B-Instruct~\citep{qwen3vl}, InternVL3.5-4B-Instruct~\citep{internvl3_5}, GPT-6 Luna~\citep{gpt6luna}, and Gemini 3.5 Flash-Lite~\citep{gemini3_5_flash_lite} as planners.

\begin{table}[t]
    \centering
    \small
    \caption{
    Results on 40 evaluation episodes (one per task). \textit{Top}: ideal oracle reference, \textit{middle}: VLAs with oracle plans, \textit{bottom}: planners with oracle physical execution. For Qwen3-VL-8B-Instruct, \textit{Learned} gives the results with $\pi_{0.5}$ / GR00T N1.7.
    }
    \label{tab:oracle_results}

    \setlength{\tabcolsep}{4pt}

    \begin{tabular*}{\columnwidth}{@{\extracolsep{\fill}} l c c}
    \toprule
    \be{Model} & \multicolumn{2}{c}{\be{Progress (\%)} $\uparrow$} \\
    \cmidrule(lr){2-3}
    & \be{Oracle} & \be{Learned} \\
    \midrule
    Oracle Plan + Oracle Physical Execution & \textbf{100.0} & -- \\
    \midrule
    \rowcolor{gray!20}
    \multicolumn{3}{l}{\textit{VLA Execution with Oracle Plans}} \\
    $\pi_{0.5}$         & 6.3           & 6.6  \\
    \quad + API         & \textbf{24.2} & 9.2  \\
    \addlinespace
    GR00T N1.7          & 9.2           & 7.0  \\
    \quad + API         & 17.2          & 10.0 \\
    \midrule
    \rowcolor{gray!20}
    \multicolumn{3}{l}{\textit{High-Level Planning with Oracle Physical Execution}} \\
    Qwen3-VL-4B-Instruct    & 25.0          & -- \\
    Qwen3-VL-8B-Instruct    & 34.1          & 9.6 / 10.8 \\
    InternVL3.5-4B-Instruct & 9.8           & -- \\
    GPT-6 Luna            & \textbf{57.3} & -- \\
    Gemini 3.5 Flash-Lite   & 46.7          & -- \\
    \bottomrule
\end{tabular*}
\end{table}

\subsection{Metrics}

\noindent\textbf{Task SR}. We measure the percentage of episodes that satisfy all goal conditions (Section~\ref{sec:benchmark_task_design}) of their task. The goal conditions are verified using object states and relations provided by OmniGibson and device states in the SimuHome environment. In a more fine-grained manner, we also compare each rollout against the subtasks of its oracle plan.

\noindent\textbf{Subtask SR}. We also measure the percentage of subtasks whose completion condition is met at any time during the rollout. For example, for a navigation subtask, a subtask is considered complete when the robot arrives within 0.8 m of its target. For a device-operation subtask, a subtask is considered complete when the device reaches the requested setting.

\noindent\textbf{Progress}. We measure the average percentage of subtasks completed before the first incomplete subtask in each episode's oracle plan. For fair comparison, subtasks that can be bypassed through API access, such as walking to a control panel, are excluded from Progress calculation for all methods. The completion criteria for individual subtasks are detailed in Appendix~\ref{app:metrics}.

\subsection{End-to-End Evaluation}
\label{sec:main_results}

\noindent\textbf{Overall Performance.}
Table~\ref{tab:main_results} presents end-to-end evaluation results. Although every task was completed by a human operator on its training episodes and by oracle physical execution of its oracle plan on all 40 diagnostic episodes (Table~\ref{tab:oracle_results}), none of the six VLA variants completes any of the 120 evaluation episodes (Task SR 0.0\%). 

\noindent\textbf{Why do models fail?} The low success rate stems from the two challenges in PAIR tasks, (1) \textbf{long task horizons} and (2) \textbf{fine-grained device control}. We find that realistic user queries from survey participants (Section~\ref{sec:request_survey}) require 9.9 subtasks on average, making failures in earlier subtasks consequential for subsequent ones. The low success rate of subtasks under the ``Place'' skill shows this challenge. To successfully place an object, the robot must first pick it up and, when placing it inside an appliance, open the door. However, in most of the times, models fail at these prerequisite steps, causing the ``Place'' skill to have a success rate of 1.2\%, despite being required in 117 of the 120 episodes. Also, models struggle to operate the realistic device control panels in PAIR, which are modeled after those of commercial appliances (Section~\ref{sec:smart_device}) and require fine-grained actions such as button pressing. This is consistent with prior work identifying such fine-grained interactions as challenging~\citep{apbot,robocasa365}.

\noindent\textbf{Effect of the planner and API access.}
We find that adding the planner without API access does not improve overall Subtask SR ($-0.7$ points for $\pi_{0.5}$ and $-1.3$ for GR00T N1.7). One reason is that the physical channel is limited by execution rather than planning. As shown in Section~\ref{sec:diagnostics}, even with the oracle plans, models fail to complete the tasks. Another reason is that the planner switches to a new subtask before the VLA finishes the current one. In particular, the planner never selects the continue or wait action, and 81--83\% of the instructions it replaces are replaced after just 5 s. On the other hand, the standalone VLAs take a median of 22--24 s to complete a pick. With API access, in contrast, overall Subtask SR rises by 1.8 and 4.0 points over + HL, a gain driven by device operation (0.4\% $\rightarrow$ 14.1\% and 14.5\%), which the planner can now complete without the VLA. Detailed analyses are provided in Appendix~\ref{app:planner_analysis}.

\begin{table}[t]
    \centering
    \small
    \caption{In-domain subtask evaluation from training demonstrations, without (+ HL) and with (+ HL (API)) API access. Subtask SR (\%); parentheses give the number of subtasks.}
    \label{tab:subtask_eval}
    \setlength{\tabcolsep}{4pt}
    \begin{tabular}{lcccc}
        \toprule
        & \multicolumn{2}{c}{$\pi_{0.5}$} & \multicolumn{2}{c}{GR00T N1.7} \\
        \cmidrule(lr){2-3}\cmidrule(lr){4-5}
        Subtask & + HL & + HL (API) & + HL & + HL (API) \\
        \midrule
        \rowcolor{gray!20}
        \multicolumn{5}{l}{\textit{Subtask SR (\%)}} \\
        Device Operation (78) & 11.5 & \textbf{35.9} & 10.3 & \textbf{38.5} \\
        Status Check (22)     & 68.2 & \textbf{86.4} & 72.7 & \textbf{77.3} \\
        \midrule
        All (100)             & 24.0 & \textbf{47.0} & 24.0 & \textbf{47.0} \\
        \bottomrule
    \end{tabular}
\end{table}

\subsection{In-Domain Subtask Evaluation}
\label{sec:subtask_eval}
API access substantially raises Subtask SR (from 24.0\% to 47.0\% for both VLAs; Table~\ref{tab:subtask_eval}) and more than triples the success of device operations (from 11.5\% to 35.9\% for $\pi_{0.5}$ and from 10.3\% to 38.5\% for GR00T N1.7), an effect that the end-to-end evaluation masks because failures in earlier subtasks leave many device operations unattempted (Section~\ref{sec:diagnostics}).

\begin{figure}[!t]
    \centering
    {\setlength{\fboxsep}{0pt}\setlength{\fboxrule}{1pt}%
    \fcolorbox{black!25}{white}{\includegraphics[width=\dimexpr\columnwidth-2pt\relax]{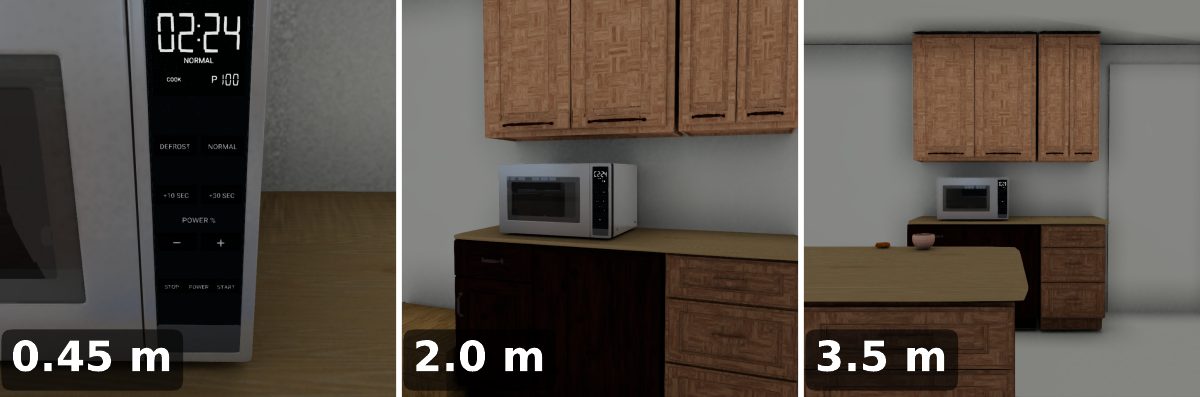}}}
    \par\vspace{9pt}
    \includegraphics[width=\columnwidth]{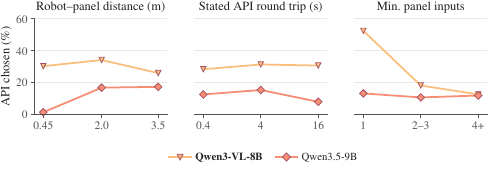}
    \caption{Top: head-camera views of one microwave panel at 0.45, 2.0 and 3.5 m. Bottom: share of device-operation cases whose first planner decision is an API command to the requested device, by distance, stated API round trip and minimum panel inputs.}
    \label{fig:panel_distances}
\end{figure}

\subsection{Oracle Evaluation}
\label{sec:diagnostics}

\begin{figure}[!t]
    \centering
    \includegraphics[width=\columnwidth]{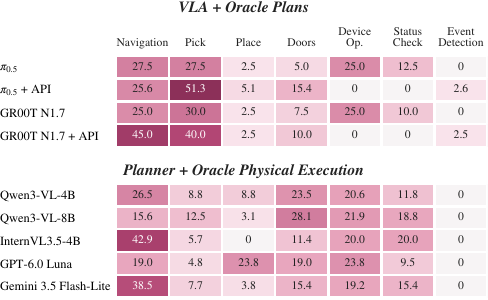}
    \caption{Skill of the first plan subtask that failed rollouts never complete (\% of each row's failed rollouts). \textit{Top}: VLAs with oracle plans. \textit{Bottom}: planners with oracle physical execution.}
    \label{fig:oracle_stop_skills}
\end{figure}

\noindent\textbf{VLA execution with oracle plans.}
Without API access, oracle plans provide little improvement over the learned planner (Table~\ref{tab:oracle_results}), suggesting that physical execution is the main bottleneck. With API access, Progress increases substantially (24.2\% vs.\ 9.2\% for $\pi_{0.5}$ and 17.2\% vs.\ 10.0\% for GR00T N1.7), mainly through correct API calls. Yet roughly three quarters of device operations remain incomplete because failures in earlier physical subtasks prevent subsequent API calls. Figure~\ref{fig:oracle_stop_skills} (up) shows the first uncompleted skill in each failed rollout. Without API access, failures are spread across navigation, picking, and device operation. With API access, they concentrate on navigation and picking (77--85\%). Only one of the 160 rollouts completes its entire plan.

\noindent\textbf{High-level planning with oracle physical execution.}
With oracle physical execution, Qwen3-VL-8B achieves 34.1\% Progress, compared with 9.6--10.8\% when paired with the VLAs. Performance varies widely across planners (9.8--57.3\%), with even the best falling well below the 100\% achieved by oracle plans. Since physical subtasks always succeed, the remaining failures reflect planning limitations. Unlike VLA execution failures, these span a broader range of skills, including navigation, device operation, door manipulation, and status checks (Figure~\ref{fig:oracle_stop_skills}, bottom). Together, these results identify physical execution and high-level planning as complementary bottlenecks.

\subsection{Channel Selection}
\label{sec:channel_selection}
To study when a planner chooses the API, we give it a single device-operation or sensor-reading request (e.g., \textit{``Select the microwave Defrost mode''}) with the robot's camera views and the prompt and tools of the main experiment, and record the channel of its first operating decision without executing it. Crossing 216 requests over all 84 device assets with three robot--panel distances (Figure~\ref{fig:panel_distances}) and three stated API round-trip delays (0.4, 4, and 16 s) yields 1,944 cases per model, 1,818 of them device operations. We report four VLMs here, GPT-6 Luna, Qwen3-VL-8B, Qwen3.5-4B, and Qwen3.5-9B, and the rest in Appendix~\ref{app:channel_selection}.

\noindent\textbf{Panel visibility matters more than API delay.}
Figure~\ref{fig:panel_distances} (bottom) shows the API share of Qwen3-VL-8B, the planner of our main experiments, and Qwen3.5-9B. Qwen3.5-9B chooses the API in only 1\% of device operations at 0.45 m but in 17\% at 2.0 and 3.5 m, and GPT-6 Luna and Qwen3.5 with reasoning shift more (5\% vs.\ 32--37\% and 19--25\% vs.\ 73--83\%), whereas Qwen3-VL-8B uses the API in 26--34\% of the cases at every distance. The stated API delay matters little: from 0.4 to 16 s, the API share of the other models moves by less than 8 points, and only GPT-6 Luna turns away from a slower API (31\% $\rightarrow$ 17\%). Neither plotted planner uses the API more for operations that need more panel inputs: Qwen3-VL-8B chooses it for 52\% of single-input operations but for only 13\% of those that need four or more inputs, and Qwen3.5-9B stays at 11--13\%.

\section{Conclusion}
\label{sec:conclusion}
PAIR couples physical appliance controls and Matter-based APIs through a shared device state and provides a survey-grounded benchmark of 40 tasks and 340 demonstrations. Neither fine-tuned VLA completes an evaluation task: API access improves device operation, but picking, placing, and door manipulation block completion. Even with oracle physical execution, the best planner reaches only 57.3\% Progress, and planners choose the API by panel visibility more than by its delay. We hope PAIR serves as a foundation for robots that act not only in the home but through it.

\section{Limitation}
\label{sec:limitation}
Our survey asked participants to write requests for PAIR's devices and objects, so the resulting tasks reflect smart-home delegation and not household activity in general, for which BEHAVIOR-1K activities~\citep{behavior1k} can be used in the same OmniGibson-based environment. We provide only 340 demonstrations (three task-level demonstrations per task), a small set for fine-tuning VLA policies that likely contributes to the low success rates in Section~\ref{sec:main_results} and reflects the cost of teleoperating a dual-arm mobile manipulator through long-horizon tasks with precise panel interaction (Appendix~\ref{app:demonstration_collection}).

\clearpage

\section*{LLM Usage Statement}
We used GPT-6.1 Terra (via the official API) for skill labeling, as described in Section~\ref{sec:request_survey}; the authors manually validated a random sample of 100 generated labels. We also utilized Claude Code and OpenAI Codex to assist with writing analysis scripts and identifying relevant literature, with the latter additionally used to help draft and format the supplementary material. Furthermore, Claude Fable 5.2 was used for grammatical editing and language polishing throughout the manuscript. All other LLM-generated outputs, including code and cited references, were thoroughly reviewed and verified by the authors, who take full responsibility for the content of this paper.

{
    \small
    \bibliographystyle{ieeenat_fullname}
    \bibliography{main}
}

\clearpage
\appendix
\onecolumn
\raggedbottom
{
\centering
\Large
\textbf{\thetitle}\\
\vspace{0.5em}Appendix \\
\vspace{1.0em}
}

\section{PAIR Details}
\label{app:pair_details}

This section describes the smart-device assets and their evaluation, and then the two channels through which an agent observes and operates them.
Both channels act on one device state, so an operation through either channel changes what the other can observe.

\subsection{Smart Devices}
\label{app:smart_devices}

Each smart-device asset combines an appliance \textit{body} with an operable control \textit{panel}.
Body variants differ in the geometry of the appliance, whereas the panel variants P1--P3 differ in how functions are selected and how feedback is shown; for example, one panel selects a program through repeated button presses and another through a knob.
All variants of a device type share a single \textit{device model} that defines the device's states, commands, and operating constraints.

The 84 assets cover 14 appliance categories and four ambient-sensor categories.
Table~\ref{tab:device_list} lists the device types and their Matter clusters, Table~\ref{tab:matter_clusters} the implemented attributes and commands, and Table~\ref{tab:asset_sources} the body sources and licenses.
The physical interface of each device type, that is, its controls, displayed state, and preconditions, is listed in Table~\ref{tab:physical_interface} (Section~\ref{app:physical_interface}).

\begingroup
\setlength{\tabcolsep}{3pt}
\renewcommand{\arraystretch}{1.2}
\setlength{\LTcapwidth}{\linewidth}

\begin{xltabular}{\linewidth}{
    >{\scriptsize\raggedright\arraybackslash}p{3.6cm}|
    >{\scriptsize\raggedright\arraybackslash}X
}
\caption{List of implemented device types and their corresponding clusters.}
\label{tab:device_list} \\

\toprule
\textbf{Device type} & \textbf{Clusters} \\
\midrule
\endfirsthead

\multicolumn{2}{c}{\footnotesize \textit{Table \thetable\ continued from previous page}} \\
\toprule
\textbf{Device type} & \textbf{Clusters} \\
\midrule
\endhead

\midrule
\multicolumn{2}{r}{\footnotesize \textit{Continued on next page}} \\
\endfoot

\bottomrule
\endlastfoot

Room Air Conditioner & Activated Carbon Filter Monitoring, Basic Information, Descriptor, Fan Control, HEPA Filter Monitoring, Identify, On/Off, Thermostat \\
\hline
Air Purifier & Activated Carbon Filter Monitoring, Basic Information, Descriptor, Fan Control, HEPA Filter Monitoring, Identify, On/Off \\
\hline
Cooktop & Basic Information, Descriptor, Identify, On/Off, Temperature Control, Temperature Measurement \\
\hline
Dimmable Light & Basic Information, Groups, Identify, Level Control, On/Off, Scenes Management \\
\hline
Dishwasher & Basic Information, Dishwasher Alarm, Dishwasher Mode, Identify, On/Off, Operational State \\
\hline
Fan & Basic Information, Fan Control, Groups, Identify, On/Off \\
\hline
Humidifier & Basic Information, Fan Control, On/Off, Relative Humidity Measurement \\
\hline
Laundry Dryer & Basic Information, Identify, Laundry Dryer Controls, Laundry Washer Mode, On/Off, Operational State \\
\hline
Laundry Washer & Basic Information, Identify, Laundry Washer Controls, Laundry Washer Mode, On/Off, Operational State \\
\hline
Microwave Oven & Basic Information, Identify, Microwave Oven Control, Microwave Oven Mode, On/Off, Operational State \\
\hline
On Off Light & Basic Information, Groups, Identify, On/Off, Scenes Management \\
\hline
Oven & Basic Information, Descriptor, Identify, On/Off, Oven Cavity Operational State, Oven Mode, Temperature Control, Temperature Measurement \\
\hline
Refrigerator & Basic Information, Descriptor, Identify, Refrigerator Alarm, RTCC Mode, Temperature Alarm, Temperature Control, Temperature Measurement \\
\hline
Window Covering Controller & Basic Information, Identify, Window Covering \\
\hline
Air Quality Sensor & Air Quality, Basic Information, Identify, PM10 Concentration Measurement \\
\hline
Humidity Sensor & Basic Information, Identify, Relative Humidity Measurement \\
\hline
Light Sensor & Basic Information, Identify, Illuminance Measurement \\
\hline
Temperature Sensor & Basic Information, Identify, Temperature Measurement \\

\end{xltabular}
\endgroup
\begingroup
\setlength{\tabcolsep}{3pt}
\renewcommand{\arraystretch}{1.2}

\newcolumntype{P}[1]{>{\scriptsize\raggedright\arraybackslash}p{#1}}

\newlength{\CwdB}
\setlength{\CwdB}{\dimexpr\textwidth - 2.8cm - 3.6cm - 6\tabcolsep\relax}

\begin{longtable}{P{2.8cm}|P{\CwdB}|P{3.6cm}}
\caption{List of implemented Matter clusters.}\label{tab:matter_clusters}\\
\hline
\textbf{Cluster} & \textbf{Attributes} & \textbf{Commands} \\
\hline
\endfirsthead

\hline
\textbf{Cluster} & \textbf{Attributes} & \textbf{Commands} \\
\hline
\endhead

\hline
\endfoot

\hline
\endlastfoot

Basic Information
& VendorName, VendorID, ProductName, ProductID
& None \\
\hline

Descriptor
& DeviceTypeList, ServerList, ClientList, PartsList, TagList
& None \\
\hline

Identify
& IdentifyTime, IdentifyType, ClusterRevision
& Identify, TriggerEffect \\
\hline

Groups
& NameSupport, ClusterRevision
& AddGroup, ViewGroup, GetGroupMembership, RemoveGroup, RemoveAllGroups, AddGroupIfIdentifying \\
\hline

Scenes Management
& SceneTableSize, FabricSceneInfo, ClusterRevision
& AddScene, ViewScene, RemoveScene, RemoveAllScenes, StoreScene, RecallScene, GetSceneMembership, CopyScene \\
\hline

On/Off
& OnOff, GlobalSceneControl, OnTime, OffWaitTime, StartUpOnOff
& Off, On, Toggle, OffWithEffect, OnWithRecallGlobalScene, OnWithTimedOff \\
\hline

Level Control
& CurrentLevel, RemainingTime, MinLevel, MaxLevel, OnOffTransitionTime, OnLevel, OnTransitionTime, OffTransitionTime, DefaultMoveRate, Options, StartUpCurrentLevel
& MoveToLevel, Move, Step, Stop, MoveToLevelWithOnOff, MoveWithOnOff, StepWithOnOff, StopWithOnOff \\
\hline

Fan Control
& FanMode, FanModeSequence, PercentSetting, PercentCurrent, WindSupport, WindSetting
& Step \\
\hline

Activated Carbon Filter Monitoring
& Condition, DegradationDirection, ChangeIndication, InPlaceIndicator
& ResetCondition \\
\hline

HEPA Filter Monitoring
& Condition, DegradationDirection, ChangeIndication, InPlaceIndicator
& ResetCondition \\
\hline

Window Covering
& Type, ConfigStatus, OperationalStatus, EndProductType, Mode, SafetyStatus, CurrentPositionLiftPercent100ths, TargetPositionLiftPercent100ths, NumberOfActuationsLift
& UpOrOpen, DownOrClose, StopMotion, GoToLiftPercentage \\
\hline

Operational State
& PhaseList, CurrentPhase, CountdownTime, OperationalStateList, OperationalState, OperationalError
& Pause, Resume, Stop, Start, OperationalCommandResponse \\
\hline

Oven Cavity Operational State
& PhaseList, CurrentPhase, CountdownTime, OperationalStateList, OperationalState, OperationalError
& Stop, Start, OperationalCommandResponse \\
\hline

Oven Mode
& SupportedModes, CurrentMode
& ChangeToMode \\
\hline

Dishwasher Mode
& SupportedModes, CurrentMode
& ChangeToMode \\
\hline

Dishwasher Alarm
& Mask, Latch, State, Supported
& Reset, ModifyEnabledAlarms \\
\hline

Laundry Washer Mode
& SupportedModes, CurrentMode
& ChangeToMode \\
\hline

Laundry Washer Controls
& SpinSpeeds, SpinSpeedCurrent, NumberOfRinses, SupportedRinses
& None \\
\hline

Laundry Dryer Controls
& SupportedDrynessLevels, SelectedDrynessLevel
& None \\
\hline

Microwave Oven Mode
& SupportedModes, CurrentMode, ClusterRevision
& None \\
\hline

Microwave Oven Control
& CookTime, MaxCookTime, PowerSetting, MinPower, MaxPower, PowerStep, ClusterRevision
& SetCookingParameters, AddMoreTime \\
\hline

RTCC Mode
& SupportedModes, CurrentMode
& ChangeToMode \\
\hline

Refrigerator Alarm
& Mask, State, Supported
& None \\
\hline

Temperature Alarm
& Mask, State, Supported, CriticalOverTemperatureThreshold, MajorOverTemperatureThreshold, MinorOverTemperatureThreshold, MinorUnderTemperatureThreshold, MajorUnderTemperatureThreshold, CriticalUnderTemperatureThreshold
& SetTemperatureAlarmThresholds \\
\hline

Temperature Control
& SelectedTemperatureLevel, SupportedTemperatureLevels, TemperatureSetpoint, MinTemperature, MaxTemperature, Step
& SetTemperature \\
\hline

Temperature Measurement
& MeasuredValue, MinMeasuredValue, MaxMeasuredValue
& None \\
\hline

Thermostat
& LocalTemperature, OccupiedCoolingSetpoint, ControlSequenceOfOperation, SystemMode, ACErrorCode
& SetpointRaiseLower \\
\hline

Relative Humidity Measurement
& MeasuredValue, MinMeasuredValue, MaxMeasuredValue
& None \\
\hline

Illuminance Measurement
& MeasuredValue, MinMeasuredValue, MaxMeasuredValue, LightSensorType
& None \\
\hline

PM10 Concentration Measurement
& MeasuredValue, MinMeasuredValue, MaxMeasuredValue, MeasurementUnit, MeasurementMedium
& None \\
\hline

Air Quality
& AirQuality
& None \\

\end{longtable}
\endgroup

\begingroup
\renewcommand{\arraystretch}{1.3}

\setlength{\LTleft}{0pt}
\setlength{\LTright}{0pt}

\begin{longtable}{@{\extracolsep{\fill}}
    >{\scriptsize\raggedright\arraybackslash}p{2.6cm}
    >{\scriptsize\raggedright\arraybackslash}p{3.2cm}
    >{\scriptsize\raggedright\arraybackslash}p{2.6cm}
    >{\scriptsize\raggedright\arraybackslash}p{2.1cm}
    >{\scriptsize\raggedright\arraybackslash}p{3.0cm}@{}
}

\caption{Sources and licenses of the smart-device body assets.}
\label{tab:asset_sources} \\

\toprule
\textbf{Device type} &
\textbf{Asset} &
\textbf{Provider} &
\textbf{Source} &
\textbf{Usage terms} \\
\midrule
\endfirsthead

\multicolumn{5}{c}{\footnotesize\textit{Table \thetable\ continued from previous page}} \\
\toprule
\textbf{Device type} &
\textbf{Asset} &
\textbf{Provider} &
\textbf{Source} &
\textbf{Usage terms} \\
\midrule
\endhead

\midrule
\multicolumn{5}{r}{\footnotesize\textit{Continued on next page}} \\
\endfoot

\bottomrule
\endlastfoot

\multirow{2}{=}{Air Purifier} &
T/Y Air Purifier &
tuma\_youzyo/T\_Y &
\href{https://sketchfab.com/3d-models/9c151de6d0f344948c73143d25ba784c}{Sketchfab} &
\href{https://creativecommons.org/licenses/by/4.0/}{CC BY 4.0} \\
&
Levoit Air Purifier &
Moryak &
\href{https://sketchfab.com/3d-models/levoit-air-purifier-7c40314b6d19433590fe26c01ffa6476}{Sketchfab} &
\href{https://creativecommons.org/licenses/by/4.0/}{CC BY 4.0} \\

\addlinespace

\multirow{2}{=}{Cooktop} &
Induction Cooker 2 Zones &
Scopia &
\href{https://sketchfab.com/3d-models/induction-cooker-2-zones-ee0fa270366d4b4cac309b9a790178e4}{Sketchfab} &
\href{https://creativecommons.org/licenses/by-sa/4.0/}{CC BY-SA 4.0} \\
&
Induction Stove &
Anshul Sharma &
\href{https://sketchfab.com/3d-models/induction-stove-28625863181849ba8a8e64000b67a952}{Sketchfab} &
\href{https://creativecommons.org/licenses/by/4.0/}{CC BY 4.0} \\

\addlinespace

Humidifier &
Tower Humidifier Concept &
MidaZ\_3D &
\href{https://sketchfab.com/3d-models/tower-humidifier-concept-b348e492942f4c32980cf8f1d3a4a13b}{Sketchfab} &
\href{https://creativecommons.org/licenses/by-nc/4.0/}{CC BY-NC 4.0} \\

\addlinespace

\multirow{3}{=}{Dishwasher} &
dishwasher\_2 &
ArtVIP &
\href{https://huggingface.co/datasets/x-humanoid-robomind/ArtVIP/tree/main/Articulated_objects/major_appliances/dishwasher/dishwasher_2}{Hugging Face} &
\href{https://www.apache.org/licenses/LICENSE-2.0}{Apache-2.0} \\
&
dishwasher\_4 &
ArtVIP &
\href{https://huggingface.co/datasets/x-humanoid-robomind/ArtVIP/tree/main/Articulated_objects/major_appliances/dishwasher/dishwasher_4}{Hugging Face} &
\href{https://www.apache.org/licenses/LICENSE-2.0}{Apache-2.0} \\
&
dishwasher\_3 &
ArtVIP &
\href{https://huggingface.co/datasets/x-humanoid-robomind/ArtVIP/tree/main/Articulated_objects/major_appliances/dishwasher/dishwasher_3}{Hugging Face} &
\href{https://www.apache.org/licenses/LICENSE-2.0}{Apache-2.0} \\

\addlinespace

\multirow{3}{=}{Laundry Washer, Laundry Dryer} &
rolling\_washer\_1 &
ArtVIP &
\href{https://huggingface.co/datasets/x-humanoid-robomind/ArtVIP/tree/main/Articulated_objects/major_appliances/washing_machine/rolling_washer_1}{Hugging Face} &
\href{https://www.apache.org/licenses/LICENSE-2.0}{Apache-2.0} \\
&
rolling\_washer\_5 &
ArtVIP &
\href{https://huggingface.co/datasets/x-humanoid-robomind/ArtVIP/tree/main/Articulated_objects/major_appliances/washing_machine/rolling_washer_5}{Hugging Face} &
\href{https://www.apache.org/licenses/LICENSE-2.0}{Apache-2.0} \\
&
rolling\_washer\_6 &
ArtVIP &
\href{https://huggingface.co/datasets/x-humanoid-robomind/ArtVIP/tree/main/Articulated_objects/major_appliances/washing_machine/rolling_washer_6}{Hugging Face} &
\href{https://www.apache.org/licenses/LICENSE-2.0}{Apache-2.0} \\

\addlinespace

\multirow{2}{=}{Oven} &
oven\_7 &
ArtVIP &
\href{https://huggingface.co/datasets/x-humanoid-robomind/ArtVIP/tree/main/Articulated_objects/major_appliances/built-in\%20oven/oven_7}{Hugging Face} &
\href{https://www.apache.org/licenses/LICENSE-2.0}{Apache-2.0} \\
&
oven\_8 &
ArtVIP &
\href{https://huggingface.co/datasets/x-humanoid-robomind/ArtVIP/tree/main/Articulated_objects/major_appliances/built-in\%20oven/oven_8}{Hugging Face} &
\href{https://www.apache.org/licenses/LICENSE-2.0}{Apache-2.0} \\

\addlinespace

\multirow{3}{=}{Microwave Oven} &
microwave\_1 &
ArtVIP &
\href{https://huggingface.co/datasets/x-humanoid-robomind/ArtVIP/tree/main/Articulated_objects/small_appliances/microwave/microwave_1}{Hugging Face} &
\href{https://www.apache.org/licenses/LICENSE-2.0}{Apache-2.0} \\
&
microwave\_2 &
ArtVIP &
\href{https://huggingface.co/datasets/x-humanoid-robomind/ArtVIP/tree/main/Articulated_objects/small_appliances/microwave/microwave_2}{Hugging Face} &
\href{https://www.apache.org/licenses/LICENSE-2.0}{Apache-2.0} \\
&
microwave\_3 &
ArtVIP &
\href{https://huggingface.co/datasets/x-humanoid-robomind/ArtVIP/tree/main/Articulated_objects/small_appliances/microwave/microwave_3}{Hugging Face} &
\href{https://www.apache.org/licenses/LICENSE-2.0}{Apache-2.0} \\

\addlinespace

Refrigerator &
fridge\_3 &
ArtVIP &
\href{https://huggingface.co/datasets/x-humanoid-robomind/ArtVIP/tree/main/Articulated_objects/major_appliances/refrigerator/fridge_3}{Hugging Face} &
\href{https://www.apache.org/licenses/LICENSE-2.0}{Apache-2.0} \\

\addlinespace

Air Conditioner &
air\_conditioning001 &
ArtVIP &
\href{https://huggingface.co/datasets/x-humanoid-robomind/ArtVIP/tree/main/Articulated_objects/major_appliances/standing_air_conditioner/air\%20conditioning001}{Hugging Face} &
\href{https://www.apache.org/licenses/LICENSE-2.0}{Apache-2.0} \\

\addlinespace

Fan &
big\_electric\_fan\_3 &
ArtVIP &
\href{https://huggingface.co/datasets/x-humanoid-robomind/ArtVIP/tree/main/Articulated_objects/major_appliances/standing\%20fan/big_electric_fan_3}{Hugging Face} &
\href{https://www.apache.org/licenses/LICENSE-2.0}{Apache-2.0} \\

\addlinespace

Window Covering &
rdacud &
Stanford Vision and Learning Lab &
\href{https://behavior.stanford.edu/knowledgebase/objects/rdacud.html}{BEHAVIOR-1K} &
\href{https://behavior.stanford.edu/}{BEHAVIOR Data Bundle EULA} \\

\addlinespace

Window Covering, Dimmable Light, On/Off Light &
Wall controller panels &
PAIR (this work) &
This work &
\href{https://www.apache.org/licenses/LICENSE-2.0}{Apache-2.0} \\

\addlinespace

Light Sensor &
\multirow{4}{=}{\makecell[l]{SEN-P1\\SEN-P2\\SEN-P3}} &
\multirow{4}{=}{PAIR (this work)} &
\multirow{4}{=}{This work} &
\multirow{4}{=}{\href{https://www.apache.org/licenses/LICENSE-2.0}{Apache-2.0}} \\*

Temperature Sensor &
& & & \\*

Humidity Sensor &
& & & \\*

Air Quality Sensor &
& & & \\

\end{longtable}

\endgroup

\paragraph{Panel dimensions.}
We compare the dimensions of PAIR's panel elements with measurements of commercial appliances (Table~\ref{tab:panel_dimensions}).
A button face is the surface that is pressed, and control spacing is the center-to-center distance between a button and its nearest neighbor.
The PAIR values are taken from the asset geometry of the 51 device packages that have a panel manifest, before any placement scaling.
PAIR buttons, their spacing, and knobs are larger than the commercial medians (16.7 vs.\ 12.9 mm for the short side of a button face, 31.2 vs.\ 18.9 mm for control spacing, and 49.2 vs.\ 38.0 mm for knob diameter), while display envelopes are similar (70.1 $\times$ 32.1 vs.\ 68.9 $\times$ 32.0 mm).

\begin{table}[H]
  \centering
  \small
  \caption{Panel element dimensions in PAIR and on commercial appliances (mm). $n$ counts measured elements and IQR is the interquartile range. PAIR values are asset dimensions before placement scaling, and PAIR spacing is the center distance from each button to its nearest button. Commercial digit, text, and icon heights are separated by element name, so that split is approximate.}
  \label{tab:panel_dimensions}
  \begin{tabular}{l rrcc rrcc}
    \toprule
    & \multicolumn{4}{c}{PAIR} & \multicolumn{4}{c}{Commercial appliances} \\
    \cmidrule(lr){2-5}\cmidrule(lr){6-9}
    Element & $n$ & Median & IQR & Range & $n$ & Median & IQR & Range \\
    \midrule
    Button face, short side & 300 & 16.7 & 13.5--21.0 & 5.0--49.6 & 102 & 12.9 & 10.0--16.3 & 6.0--43.3 \\
    Control spacing & 300 & 31.2 & 25.1--41.9 & 13.0--227.7 & 363 & 18.9 & 16.6--20.5 & 2.6--216.5 \\
    Knob diameter & 10 & 49.2 & 43.4--62.9 & 33.4--71.0 & 11 & 38.0 & 27.5--42.0 & 26.5--44.5 \\
    Display width & 18 & 70.1 & 43.6--80.7 & 38.0--107.6 & 21 & 68.9 & 47.5--95.7 & 32.9--209.2 \\
    Display height & 18 & 32.1 & 11.1--55.9 & 7.6--65.0 & 21 & 32.0 & 20.0--55.4 & 12.0--92.0 \\
    Digit height & 123 & 11.9 & 9.2--18.7 & 7.6--30.0 & 168 & 5.5 & 3.1--10.4 & 1.0--62.4 \\
    Label text height & 273 & 3.9 & 3.3--5.0 & 1.4--50.6 & 221 & 5.6 & 4.2--8.6 & 0.8--402.7 \\
    Printed icon height & 10 & 9.4 & 8.0--10.9 & 5.9--14.0 & 336 & 7.2 & 4.0--10.5 & 0.3--61.0 \\
    \bottomrule
  \end{tabular}
\end{table}

\subsection{Panel Realism Evaluation}
\label{app:panel_realism}

\paragraph{Procedure.}
Raters judged whether each panel could plausibly appear on a contemporary commercial appliance of the same type.
Each item showed a whole-device image for context and a close-up of the panel, both in the same operating state.
Raters were asked to score only what was visible on the controls, display, labels, and panel surface; for sensors without controls, they scored the display and the surrounding face.
The instructions asked them to judge realism rather than personal preference, expected usability, or similarity to a particular brand.

\paragraph{Criteria.}
Each panel received three five-point scores.
\textit{Appearance} covers the shapes, colors, materials, finishes, and rendering of individual elements.
\textit{Layout} covers their placement, grouping, spacing, alignment, and relative size, including the display content.
\textit{Information} covers whether labels, icons, modes, values, units, and indicators suit the device and agree with one another.
Raters assigned each issue to a single criterion and did not penalize features that were not shown.
A score of 5 means that nothing looks unrealistic, 3 means clear flaws in an otherwise plausible panel, and 1 means a largely implausible panel.

\paragraph{Coverage and results.}
Fifteen raters scored 90 panels, with each rater scoring 30 panels and each panel receiving five ratings.
Six of these panels belong to two refrigerator bodies that are not among the final 84 assets, so Table~\ref{tab:panel_realism_scores} reports the remaining 84 assets with 420 ratings per criterion.
Averaged over the three criteria, the realism score is 4.63/5.
The study measures visual plausibility from static images only.

\begin{table}[H]
  \centering
  \caption{Panel realism on the 84 assets. Each criterion has 420 scores from five raters per panel. Standard deviations are over individual ratings.}
  \label{tab:panel_realism_scores}
  \begin{tabular}{lcc}
    \toprule
    Criterion & Mean & Standard deviation \\
    \midrule
    Appearance & 4.67 & 0.62 \\
    Layout & 4.48 & 0.77 \\
    Information & 4.76 & 0.53 \\
    \bottomrule
  \end{tabular}
\end{table}

\subsection{Agent Interface}
\label{app:agent_interface}

An agent can operate a device either physically or through the API.
Physical control requires the robot to reach, identify, and actuate a control, whereas API control requires the agent to discover the device and name a valid endpoint, cluster, and attribute or command.
Both channels are subject to the same device preconditions, such as a closed door before a cycle can start.
The API exposes only device and room state, so object poses and grasp states have to be observed through the cameras.

\subsubsection{Physical Interface}
\label{app:physical_interface}

Pressing a button or turning a knob to a detent produces a \textit{physical input}, which the device applies to its state before updating its displays and indicators.
Physical preconditions, such as a closed door or a loaded object, are read from the scene whenever an operation is attempted, and they apply to API commands in the same way.
Table~\ref{tab:physical_interface} lists the inputs, displayed state, and preconditions of each device type.

\begingroup
\renewcommand{\arraystretch}{1.3}
\setlength{\LTcapwidth}{\linewidth}

\begin{xltabular}{\linewidth}{
    >{\scriptsize\hsize=0.62\hsize\raggedright\arraybackslash}X
    >{\scriptsize\hsize=1.15\hsize\raggedright\arraybackslash}X
    >{\scriptsize\hsize=1.43\hsize\raggedright\arraybackslash}X
    >{\scriptsize\hsize=0.80\hsize\raggedright\arraybackslash}X
}
\caption{Physical triggers, observables and preconditions of each device type. Triggers are the union over the panel variants of a device type.}
\label{tab:physical_interface} \\
\toprule
\textbf{Device type}
& \textbf{Physical trigger}
& \textbf{Physical observable}
& \textbf{Physical precondition} \\
\midrule
\endfirsthead

\multicolumn{4}{c}{\footnotesize \textit{Table \thetable\ continued from previous page}} \\
\toprule
\textbf{Device type} & \textbf{Physical trigger} & \textbf{Physical observable} & \textbf{Physical precondition} \\
\midrule
\endhead

\midrule
\multicolumn{4}{r}{\footnotesize \textit{Continued on next page}} \\
\endfoot

\bottomrule
\endlastfoot

Room Air Conditioner & Wall controller: power button, temperature up button, temperature down button, fan button & On\allowbreak Off.\allowbreak On\allowbreak Off, Thermostat.\allowbreak Occupied\allowbreak Cooling\allowbreak Setpoint, Thermostat.\allowbreak Local\allowbreak Temperature, Fan\allowbreak Control.\allowbreak Fan\allowbreak Mode & None \\
\addlinespace
Air Purifier & On-body panel: power button, fan button & On\allowbreak Off.\allowbreak On\allowbreak Off, Fan\allowbreak Control.\allowbreak Fan\allowbreak Mode, Fan\allowbreak Control.\allowbreak Percent\allowbreak Current & None \\
\addlinespace
Cooktop & On-body panel: power button, zone power buttons, zone select buttons, temperature up button, temperature down button, heat-level knob, heat-level sliders & ep2 On\allowbreak Off.\allowbreak On\allowbreak Off, ep2 Temperature\allowbreak Control.\allowbreak Selected\allowbreak Temperature\allowbreak Level, ep3 On\allowbreak Off.\allowbreak On\allowbreak Off, ep3 Temperature\allowbreak Control.\allowbreak Selected\allowbreak Temperature\allowbreak Level & None \\
\addlinespace
Dimmable Light & Wall controller: power button, brightness up button, brightness down button & On\allowbreak Off.\allowbreak On\allowbreak Off, Level\allowbreak Control.\allowbreak Current\allowbreak Level & None \\
\addlinespace
Dishwasher & On-body panel: power button, start/pause/resume button, stop button, program buttons, mode button & On\allowbreak Off.\allowbreak On\allowbreak Off, Dishwasher\allowbreak Mode.\allowbreak Current\allowbreak Mode, Dishwasher\allowbreak Alarm.\allowbreak State, Operational\allowbreak State.\allowbreak Operational\allowbreak State, Operational\allowbreak State.\allowbreak Current\allowbreak Phase, Operational\allowbreak State.\allowbreak Countdown\allowbreak Time, Operational\allowbreak State.\allowbreak Operational\allowbreak Error & Door closed, Load inside, Obstruction clear \\
\addlinespace
Fan & On-body panel: power button, speed button, mode button & On\allowbreak Off.\allowbreak On\allowbreak Off, Fan\allowbreak Control.\allowbreak Fan\allowbreak Mode, Fan\allowbreak Control.\allowbreak Wind\allowbreak Setting & None \\
\addlinespace
Humidifier & On-body panel: power button, fan button & On\allowbreak Off.\allowbreak On\allowbreak Off, Fan\allowbreak Control.\allowbreak Percent\allowbreak Current, ep2 Relative\allowbreak Humidity\allowbreak Measurement.\allowbreak Measured\allowbreak Value & None \\
\addlinespace
Laundry Dryer & On-body panel: power button, start button, pause/resume button, stop button, mode button, dryness button, dryness knob, program knob & On\allowbreak Off.\allowbreak On\allowbreak Off, Laundry\allowbreak Washer\allowbreak Mode.\allowbreak Current\allowbreak Mode, Laundry\allowbreak Dryer\allowbreak Controls.\allowbreak Selected\allowbreak Dryness\allowbreak Level, Operational\allowbreak State.\allowbreak Operational\allowbreak State, Operational\allowbreak State.\allowbreak Countdown\allowbreak Time, Operational\allowbreak State.\allowbreak Operational\allowbreak Error & Door closed, Load inside \\
\addlinespace
Laundry Washer & On-body panel: power button, start/pause/resume button, stop button, spin button, rinse button, mode button, program knob & On\allowbreak Off.\allowbreak On\allowbreak Off, Laundry\allowbreak Washer\allowbreak Mode.\allowbreak Current\allowbreak Mode, Laundry\allowbreak Washer\allowbreak Controls.\allowbreak Spin\allowbreak Speed\allowbreak Current, Laundry\allowbreak Washer\allowbreak Controls.\allowbreak Number\allowbreak Of\allowbreak Rinses, Operational\allowbreak State.\allowbreak Operational\allowbreak State, Operational\allowbreak State.\allowbreak Current\allowbreak Phase, Operational\allowbreak State.\allowbreak Countdown\allowbreak Time, Operational\allowbreak State.\allowbreak Operational\allowbreak Error & Door closed, Load inside, Obstruction clear \\
\addlinespace
Microwave Oven & On-body panel: power button, start button, stop button, normal button, defrost button, power-level up button, power-level down button, add-time buttons, door-release switch & On\allowbreak Off.\allowbreak On\allowbreak Off, Microwave\allowbreak Oven\allowbreak Mode.\allowbreak Current\allowbreak Mode, Microwave\allowbreak Oven\allowbreak Control.\allowbreak Cook\allowbreak Time, Microwave\allowbreak Oven\allowbreak Control.\allowbreak Power\allowbreak Setting, Operational\allowbreak State.\allowbreak Operational\allowbreak State, Operational\allowbreak State.\allowbreak Current\allowbreak Phase, Operational\allowbreak State.\allowbreak Countdown\allowbreak Time, Operational\allowbreak State.\allowbreak Operational\allowbreak Error & Door closed, Load inside, Obstruction clear \\
\addlinespace
On Off Light & Wall switch panel: power button & On\allowbreak Off.\allowbreak On\allowbreak Off & None \\
\addlinespace
Oven & On-body panel: power button, start button, stop button, temperature up button, temperature down button, mode buttons & On\allowbreak Off.\allowbreak On\allowbreak Off, ep2 Oven\allowbreak Mode.\allowbreak Current\allowbreak Mode, ep2 Oven\allowbreak Cavity\allowbreak Operational\allowbreak State.\allowbreak Operational\allowbreak State, ep2 Oven\allowbreak Cavity\allowbreak Operational\allowbreak State.\allowbreak Current\allowbreak Phase, ep2 Oven\allowbreak Cavity\allowbreak Operational\allowbreak State.\allowbreak Countdown\allowbreak Time, ep2 Oven\allowbreak Cavity\allowbreak Operational\allowbreak State.\allowbreak Operational\allowbreak Error, ep2 Temperature\allowbreak Control.\allowbreak Temperature\allowbreak Setpoint & Door closed, Load inside, Obstruction clear \\
\addlinespace
Refrigerator & On-door panel: temperature up button, temperature down button, temperature button, mode buttons & RTCCMode.\allowbreak Current\allowbreak Mode, Temperature\allowbreak Control.\allowbreak Temperature\allowbreak Setpoint, Temperature\allowbreak Measurement.\allowbreak Measured\allowbreak Value & Door closed to increase cooling \\
\addlinespace
Window Covering Controller & Wall controller: open button, stop button, close button & Window\allowbreak Covering.\allowbreak Safety\allowbreak Status & None \\
\addlinespace
Air Quality Sensor & None & PM10\allowbreak Concentration\allowbreak Measurement.\allowbreak Measured\allowbreak Value & None \\
\addlinespace
Humidity Sensor & None & Relative\allowbreak Humidity\allowbreak Measurement.\allowbreak Measured\allowbreak Value & None \\
\addlinespace
Light Sensor & None & Illuminance\allowbreak Measurement.\allowbreak Measured\allowbreak Value & None \\
\addlinespace
Temperature Sensor & None & Temperature\allowbreak Measurement.\allowbreak Measured\allowbreak Value & None \\
\end{xltabular}
\endgroup

\subsubsection{List of Tools}
\label{app:list_of_tools}

The API tools cover device discovery, documentation, state reads, attribute writes, and device commands (Table~\ref{tab:tool_list}).
A typical agent first finds the devices in a room, inspects a device's structure, and consults the cluster documentation before issuing an operation.
Returned values and errors are added to the agent's history, and devices must be addressed by the identifiers that discovery returns.

\begin{table}[H]
\caption{Tool List for Agent}
\label{tab:tool_list}
\centering
{
\scriptsize
\renewcommand{\arraystretch}{1.2}
\begin{tabularx}{\linewidth}{l X X}
\toprule
\textbf{Name} & \textbf{Description} & \textbf{Args} \\
\midrule
execute\_command & Execute a command on a device (e.g., turn on light, set level, set setpoint). & \texttt{device\_id} (str, req); \texttt{endpoint\_id} (int, req); \texttt{cluster\_id} (str, req); \texttt{device\_type} (str, req); \texttt{command\_id} (str, req); \texttt{args} (dict, req). \\
\addlinespace
write\_attribute & Directly set a device attribute value. & \texttt{device\_id} (str, req); \texttt{endpoint\_id} (int, req); \texttt{cluster\_id} (str, req); \texttt{device\_type} (str, req); \texttt{attribute\_id} (str, req); \texttt{value} (any, req). \\
\addlinespace
get\_all\_attributes & Get all attributes of a device. & \texttt{device\_id} (str, req). \\
\addlinespace
get\_attribute & Get a specific attribute of a device. & \texttt{device\_id} (str, req); \texttt{endpoint\_id} (int, req); \texttt{cluster\_id} (str, req); \texttt{attribute\_id} (str, req). \\
\addlinespace
get\_device\_structure & Get device structure (endpoints, clusters, attributes, and commands). & \texttt{device\_id} (str, req). \\
\addlinespace
get\_rooms & Get all rooms in the home along with their display names. & (none) \\
\addlinespace
get\_room\_devices & Get all devices in a room. & \texttt{room\_id} (str, req). \\
\addlinespace
get\_room\_states & Get environmental states of a room (temperature, humidity, illuminance, PM10). & \texttt{room\_id} (str, req). \\
\addlinespace
get\_cluster\_doc & Get the public Matter documentation of a cluster (its attributes, commands and dependencies). & \texttt{cluster\_id} (str, req). \\
\addlinespace
get\_environment\_control\_rules & Get control rules for a specific environmental state. & \texttt{state} (str, req). \\
\bottomrule
\end{tabularx}
}
\end{table}

\subsubsection{Simulation Time and API Delay}
\label{app:simulation_time}

OmniGibson and SimuHome share one simulation clock: after each robot action frame, the device model advances by the same interval and updates its feedback.
An API request is subject to a configurable one-way delay $d$, so it reads or changes the device state after $d$ and its response arrives after $2d$, with preconditions checked at the time the operation executes.
Physical control involves no network delay, and the clock is paused during model inference.

\section{Survey Details}
\label{app:survey_details}

We surveyed what people would ask a household robot to do in a smart home.
Participants wrote requests in their own words and marked the objects and device functions that each request involved.

\subsection{Survey Questionnaire}
\label{app:survey_questionnaire}

The survey first introduced the robot's appearance and capabilities: navigating the home, handling objects, operating appliance controls, and remotely accessing smart devices (Figure~\ref{fig:survey_meet_the_robot}).
Each item then showed a household scene together with the functions of the available devices, drawn from 14 appliance categories, four sensor categories, and 124 household objects, and asked \textit{``What would you ask the robot to do?''}

The survey had a single-device part and a multi-device part.
In a single-device item, the participant wrote a request for an assigned device, selected the household objects involved, and marked the requested device operations (power, operating modes, numerical settings, status checks, and cycle operations such as start and stop).
In a multi-device item, the participant chose two or more devices and wrote a request that used them together.
Figures~\ref{fig:survey_questionnaire} and~\ref{fig:survey_questionnaire_requests} show the interface.


\subsection{Survey Filtering}
\label{app:survey_filtering}
We kept a request if it described an interpretable task that operates or checks the required device or devices.
Requests were excluded if they were empty, did not operate or check the assigned device, involved fewer than two devices in a multi-device item, or could not be interpreted from their text and selections.
Merely naming an appliance or using it as a support surface did not count as operating it; moving an object from a cooktop to a dishwasher, for example, operates neither appliance.

Of the 1,644 collected requests, 201 were excluded under one of these four criteria, leaving 1,443 (Table~\ref{tab:query_filtering}).
The retained requests include some that PAIR cannot simulate.

\begin{table}[H]
  \centering
  \small
  \caption{Survey request filtering. Of the 1,644 requests collected,
  201 were excluded on the basis of four criteria, and the remaining
  1,443 were retained for analysis. Each excluded request is counted
  under a single criterion.}
  \label{tab:query_filtering}
  \begin{tabularx}{\textwidth}{lXXr}
    \toprule
    Filtering criterion & Definition & Example & Requests \\
    \midrule
    Collected requests & & & 1,644 \\
    \midrule
    Non-response & No task request was provided. &
    \emph{``No objects needed.''} & 2 \\
    \addlinespace
    Device not in use & The request did not ask for operation or a status check of the assigned device. &
    \emph{``Write the word `John' on the air purifier with the pen''} & 159 \\
    \addlinespace
    Multi-device requirement & The request did not ask for operation or status checks of at least two devices in a multi-device item. &
    \emph{``There is a bowl on the cooktop, please put it in the dishwasher''} & 34 \\
    \addlinespace
    Uninterpretable response & The task, target, or operating condition could not be reliably determined from the request and its context. &
    \emph{``Turn the ac on the temperature sensor''} & 6 \\
    \midrule
    Retained requests & & & 1,443 \\
    \bottomrule
  \end{tabularx}
\end{table}

\subsection{Survey Statistics}
\label{app:survey_statistics}

The retained data contain 1,443 requests from 108 participants.
We call a request \textit{single-device} if its selections involve one device type, counting sensors, and \textit{multi-device} otherwise; 1,169 requests are single-device and 274 are multi-device.

Figure~\ref{fig:survey_stats} reports two kinds of skill counts.
The skill-distribution chart counts each skill type at most once per request and panel variant, whereas the skills-per-request chart counts every repeated physical action, such as pressing buttons for two settings, and each perception skill once.
In both charts, panel variants are weighted equally within a request and requests are weighted equally overall.
Section~\ref{app:skill_mapping} describes how device functions are expanded into physical skills.

At the request level, 842 requests (58.4\%) carry a Status Check or Event Detection label.
If every device operation and device-state observation were available through an API, 572 requests (39.6\%) could be completed through APIs alone, while 871 (60.4\%) would still need the robot's physical actions or perception.
This classification assumes a general device API and does not depend on Matter support.
Finally, 130 requests ask the robot to observe an ambient value, either unconditionally or as the trigger for another action.

\subsection{Survey Ethics}
We paid participants £9 per hour (about \$12) and balanced them by age group (20--29, 30--39, 40--49, 50--59, 60--100) and gender.
Participants who reported technical errors through Prolific were paid in full even when the survey ended early, and we fixed the reported errors. We got consent from all survey participants.

\section{Skill Taxonomy and Analysis}
\label{app:skills_taxonomy}
We describe the capabilities that the requests require with a taxonomy of robot skills.
The taxonomy captures what a request asks the robot to do, while a separate mapping (Section~\ref{app:skill_mapping}) determines which physical skills a device function requires on a given panel.

\subsection{Skill Taxonomy Construction}

Skills fall into three categories: \textit{Physical} skills change the world, \textit{Perception} skills acquire information, and \textit{Human Interaction} skills exchange information with a person.
We took the physical skills from RoboCasa~\citep{robocasa} and the category definitions from \citet{skill_survey}, and added skills that the requests needed but neither source covered.
Table~\ref{tab:full_skills_taxonomy} gives the definitions and examples, and an additional \textit{Other} label marks requested work outside the 18 skills.

\paragraph{Labeling unit.}
We labeled the skills that a request or its device-function selections express, without adding prerequisites that a robot would have to infer.
Objects, destinations, modes, and numerical values are treated as parameters of a skill.
A request to set a device therefore receives Device Operation whether or not it names a button or knob.
Status Check refers to reading a state and Event Detection to observing until a stated change occurs, so a condition stated in a request is labeled as an observation.

\paragraph{LLM labeling.}
GPT-6.1 Terra received the request, the survey instructions, the participant's selections, the mode descriptions, and the skill definitions, but not the device-function-to-skill mapping.
For each label it returned a skill code, a supporting quotation from the request, and a reason, and we checked every quotation against the request text.
For the agreement analysis, repeated occurrences of a skill within one request count once.

\begin{table}[H]
\centering
\caption{Skills by category. Objects, parts, locations, modes and values are parameters of each skill.}
\label{tab:full_skills_taxonomy}
\footnotesize
\setlength{\tabcolsep}{4pt}
\renewcommand{\arraystretch}{1.1}

\begin{tabularx}{\textwidth}{
    >{\raggedright\arraybackslash}p{52pt}
    >{\raggedright\arraybackslash}X
    l
    l
}
\toprule
\textbf{Category} & \textbf{Skill} & \textbf{Definition} & \textbf{Example} \\
\midrule
\multirow{11}{=}{\textbf{Physical}}
& Pick and Place & Grasp an object and place it at a target. & ``Put the shirt in the washer.'' \\
\cmidrule{2-4}
& Opening and Closing Doors & Open or close a hinged or sliding door. & ``Close the dishwasher door.'' \\
\cmidrule{2-4}
& Twisting Knobs & Rotate a knob or dial to a target position. & ``Turn the stove knob to high.'' \\
\cmidrule{2-4}
& Pressing Buttons & Press a button or switch. & ``Press the fan's power button.'' \\
\cmidrule{2-4}
& Insertion and Removal & Seat a held object in a receptacle, or remove it. & ``Plug in the humidifier.'' \\
\cmidrule{2-4}
& Navigation & Move the base to a location or person. & ``Go to the kitchen.'' \\
\cmidrule{2-4}
& Deformable Object Manipulation & Fold, unfold or spread a deformable object. & ``Fold the bath towels.'' \\
\cmidrule{2-4}
& Wiping and Sweeping & Wipe, scrub, mop or sweep a surface. & ``Wipe the table.'' \\
\cmidrule{2-4}
& Cooking Tool Use & Cut, peel, mix, stir or flip food with a held utensil. & ``Chop the strawberries.'' \\
\cmidrule{2-4}
& Pouring and Scooping & Transfer liquid or granules into a target. & ``Water the pot plant.'' \\
\cmidrule{2-4}
& Device Operation & Set a device's power, mode, setting or operating cycle. & ``Set the fan to natural mode.'' \\
\midrule

\multirow{5}{=}{\textbf{Perception}}
& Status Check & Read a device's state or a sensor's value. & ``Is the dryer still running?'' \\
\cmidrule{2-4}
& Object Search and Identification & Find or identify a described entity. & ``Find my scarf.'' \\
\cmidrule{2-4}
& Object State Estimation & Judge a non-pose attribute of an object. & ``Check if the towels are dry.'' \\
\cmidrule{2-4}
& Human State and Activity Recognition & Determine a person's presence or activity. & ``Dim the light if I'm asleep.'' \\
\cmidrule{2-4}
& Event Detection & Observe until a stated change occurs. & ``When the dryer finishes, \ldots'' \\
\midrule

\multirow{2}{=}{\textbf{Human}\newline\textbf{Interaction}}
& Spoken Cue Recognition & Recognize a spoken cue during the task. & ``Stop when I say so.'' \\
\cmidrule{2-4}
& Informing & Tell a person a state, result or answer. & ``Tell me the humidity.'' \\

\bottomrule
\end{tabularx}
\end{table}

\subsection{Skill Labeling Validation}
\label{app:label_validation}

We validated the LLM labels on 100 requests: 40 drawn at random, 55 selected to cover rare skills, and five more drawn at random (Table~\ref{tab:label_validation_skill_distribution}).
Because rare skills are oversampled, the agreement on this sample should not be read as an estimate over all requests.
Three authors annotated each request independently using the taxonomy and labeling guidelines; they saw the request text and the participant's selections but not the LLM labels.
Device Operation and Status Check labels implied by the participant's selections were fixed in the annotation interface, and the majority vote of the three annotators served as the reference label for each of the 18 skills and Other.

\begin{table}[H]
  \caption{Skill frequencies in the 100 requests used for labeling validation.}
  \label{tab:label_validation_skill_distribution}
  \centering
  \small
  \begin{tabular}{l r @{\hspace{20pt}} l r}
    \toprule
    \textbf{Skills} & \textbf{Requests} &
    \textbf{Skills} & \textbf{Requests} \\
    \midrule
    Device Operation               & 99 & Cooking Tool Use                     & 10 \\
    Status Check                   & 57 & Pouring and Scooping                 & 10 \\
    Pick and Place                 & 47 & Object Search and Identification     & 10 \\
    Event Detection                & 13 & Object State Estimation              & 10 \\
    Insertion and Removal          & 12 & Informing                            & 10 \\
    Navigation                     & 12 & Other                                & 10 \\
    Opening and Closing Doors      & 10 & Human State and Activity Recognition &  9 \\
    Pressing Buttons               & 10 & Spoken Cue Recognition               &  4 \\
    Deformable Object Manipulation & 10 & Twisting Knobs                       &  1 \\
    Wiping and Sweeping            & 10 &                                      &    \\
    \bottomrule
  \end{tabular}
\end{table}

We compared the LLM labels with the majority vote using micro-averaged F1 and pooled binary Cohen's $\kappa$ over 1,900 request--skill decisions (100 requests $\times$ 19 labels), and measured human agreement with the same metrics averaged over the three annotator pairs.
The LLM reached an F1 of 0.881 and a $\kappa$ of 0.856 against the majority vote, compared with a mean pairwise human agreement of 0.827 and 0.791 (Table~\ref{tab:label_agreement}).

\begin{table}[h]
  \centering
  \small
  \caption{Skill-labeling agreement on 100 requests. Human--human results are averaged across three annotator pairs.}
  \label{tab:label_agreement}
  \begin{tabular}{lcc}
    \toprule
    Comparison & F1 & $\kappa$ \\
    \midrule
    LLM vs.\ human majority & .881 & .856 \\
    Human vs.\ human & .827 & .791 \\
    \bottomrule
  \end{tabular}
\end{table}

\subsection{Device-Function-to-Skill Mapping}
\label{app:skill_mapping}

A Device Operation label does not say which physical skills a particular panel requires.
We therefore map each selected device function to the controls and feedback of every body and panel variant that supports it.
Selecting a program, for example, requires Pressing Buttons on one panel and Twisting Knobs on another, and reading the resulting feedback adds Status Check.
Skills unrelated to device operation, such as moving a shirt, keep their original labels.

For a request that involves several devices, a \textit{variant combination} fixes one body and panel for each device.
We expand the device functions within each combination and merge the resulting skills with the request's other labels.
In the skill distribution, a skill counts once per combination, and the combinations are averaged within the request.
In the skills-per-request count, every requested operation adds its physical skills, so the same skill can be counted several times.

One request, which asks to switch a refrigerator off, has no physical method on any panel and contributes only its other skills.

\section{Benchmark Details}
\label{app:benchmark_details}

The benchmark turns survey requests into executable tasks with explicit initial states and simulator-verifiable goals.
Training and evaluation share the 40 tasks and use disjoint episodes, so evaluation measures execution on new instances of known tasks.

\subsection{Task Selection}
\label{app:task_filtering}
We started from the 871 requests that combine a device operation or observation with a step that no API can perform.
We first removed 327 requests that cannot be executed in the simulator: 290 need objects, device functions, goal conditions, or skills that PAIR does not support, 10 ask for a device operation that PAIR offers through neither channel, and 27 require moving a device that is fixed in place.
Basic filtering then removed 148 more: 38 contain no object-manipulation step, and 110 require handing an object to a person, notifying the user, or wiping a surface.
The remaining 396 requests can be solved by physical actions alone and cover navigation, object transport, door operation, device control, status observation, and waiting for device or environmental events.
From these we selected 40 tasks that span the supported devices and skill combinations, ten of which involve several devices.
The selection aims for coverage rather than matching how often each kind of request occurs.
Table~\ref{tab:represent_tasks_stats} summarizes the skills and devices of the selected tasks.

\subsection{Episode Instantiation}
\label{app:episode_instantiation}
A task fixes an instruction and a goal specification, and an \textit{episode} binds them to concrete objects, devices, placements, a robot base pose, and initial device and ambient states.
We randomize these initial conditions within the limits set by the request, so that any starting relation it names holds and any environmental threshold is reachable within the episode budget.

Each task has six episodes, three for training and three for evaluation, which gives 120 episodes per split.
Because an episode also fixes the body and panel of each device, all methods face the same interface and initial state.
Together, the 120 evaluation episodes use 54 of the 84 assets and cover all 18 device types.
Each episode comes with an \textit{oracle plan}, the sequence of subtasks that completes the task from its initial state.
The plan is generated by instantiating the task's subtask template for that initial state; for example, the door-opening subtask is omitted when the door is already open.
The oracle plans of the evaluation episodes contain 9.9 subtasks on average (3--25), or 9.2 when device operations are performed through the API.
These plans are used for scoring and diagnostics and are never shown to the learned planner.

\subsection{Demonstration Collection and Annotation}
\label{app:demonstration_collection}
For each training episode, an operator recorded one successful execution of the task by teleoperating XLeRobot's base, arms, and grippers with Joy-Con controllers and operating the appliances through their panels.
Each recording contains the head and two wrist images, proprioception, and actions, and is annotated with subtask boundaries and instructions.
We also recorded 220 \textit{atomic} demonstrations, each covering a single physical skill.
Together with the 120 task-level demonstrations, this gives 340 demonstrations and 5.7 hours of execution.
Collecting them took a single operator about 100 hours, including failed and repeated attempts.
Section~\ref{app:training_details} describes how the task-level, subtask-level, and atomic views are sampled during training.

A device-operation subtask can contain several button presses or knob movements; setting a temperature, for instance, is a single subtask no matter how many inputs it needs.
An inspection subtask, by contrast, acquires information without changing the device.

\paragraph{Demonstration statistics.}
Table~\ref{tab:demo_stats_full} reports the statistics of the task-level demonstrations.
The task-level demonstrations account for 4.35 hours and the atomic demonstrations for 1.34 hours (21.9 s each on average).
Durations count recorded action frames at 30 Hz and exclude appliance time skipped by an accelerated wait.
A task-level demonstration has 15.0 annotated subtasks on average, or 9.9 without inspections and waits.
Navigation takes 37.3\% of the task-level time, followed by Pick and Place (23.0\%), Opening and Closing Doors (16.4\%), Device Operation (13.1\%), visual verification of objects and doors (5.3\%), Status Check (3.4\%), and waiting (1.5\%).
\textit{Minimum panel inputs} is the smallest number of button presses and knob turns that reaches the device goals of a demonstration from its initial device state.

\begin{table}[H]
  \centering
  \small
  \caption{Statistics of the 120 task-level demonstrations. Manipulation covers picking, placing, door, and device-operation subtasks; status checks are inspections of a device display or control panel.}
  \label{tab:demo_stats_full}
  \begin{tabular}{lrrr}
    \toprule
    Per demonstration & Mean & Median & Range \\
    \midrule
    Duration (s) & 130.5 & 119.9 & 27.3--343.3 \\
    Action steps & 3,916 & 3,598 & 820--10,298 \\
    Annotated subtasks & 15.0 & 13.5 & 6--37 \\
    \quad excluding inspections and waits & 9.9 & 9.0 & 3--25 \\
    \quad manipulation & 6.4 & 5.0 & 2--15 \\
    \quad device operation & 2.2 & 2.0 & 0--6 \\
    \quad status checks & 2.0 & 2.0 & 0--6 \\
    Skills & 4.7 & 5.0 & 3--6 \\
    Devices & 1.4 & 1.0 & 1--4 \\
    Minimum panel inputs & 3.4 & 2.0 & 0--18 \\
    \bottomrule
  \end{tabular}
\end{table}

\section{Training Details}
\label{app:training_details}

Both policies are fine-tuned in a single stage from their released base checkpoints on the same 340 PAIR demonstrations (about 5.7 hours): 120 task-level and 220 atomic demonstrations. We split the data by source demonstration into 305 training demonstrations (108 task-level, 197 atomic) and 35 validation demonstrations (12 task-level, 23 atomic), and compute normalization statistics on the training split only. Each task-level demonstration provides a task-level view, the full trajectory with the original task instruction, and a subtask-level view, each annotated subtask interval with its subtask instruction; wait intervals are excluded. Atomic demonstrations are used whole with their own instructions.

Each training sample pairs one observation (the head and two wrist RGB images and the 17-dimensional proprioceptive state) with the next 32 actions. Action windows never cross a recording gap or a wait, and subtask-level windows never leave their interval. Each sample first draws a view with probability 0.40 (task-level), 0.45 (subtask-level), or 0.15 (atomic) and then draws a window uniformly within that view. Both policies are trained for 15{,}000 steps with a global batch of 160 (2.4M sampled windows) on 5$\times$H200 GPUs. $\pi_{0.5}$ is fully fine-tuned. For GR00T~N1.7 we follow the released post-training recipe and update the projector, the action head, and the vision--language layer norms (1.62B of 3.14B parameters) while keeping the language and vision backbones frozen. Peak learning rates are the released defaults scaled by $\sqrt{2}$ for the larger batch. Table~\ref{tab:training_hparams} lists the hyperparameters. We evaluate $\pi_{0.5}$ at step 12{,}000 and GR00T~N1.7 at step 13{,}500.

\begin{table}[h]
    \centering
    \small
    \caption{Fine-tuning hyperparameters. Both policies use a cosine schedule with linear warmup and share the training data, view mixture, batch size, and number of steps.}
    \label{tab:training_hparams}
    \begin{tabular}{lcc}
    \toprule
    Configuration & $\pi_{0.5}$ & GR00T N1.7 \\
    \midrule
    Trained parameters   & All & 1.62B of 3.14B \\
    Optimizer            & AdamW ($\beta$ = 0.9, 0.95) & AdamW ($\beta$ = 0.9, 0.999) \\
    Batch size           & 160 & 160 \\
    Training steps       & 15,000 & 15,000 \\
    Learning rate        & $3.5{\times}10^{-5} \rightarrow 3.5{\times}10^{-6}$ & $1.4{\times}10^{-4} \rightarrow 0$ \\
    Warmup steps         & 1,000 & 750 \\
    Weight decay         & 0.01 & $10^{-5}$ \\
    Gradient clipping    & 1.0 & 1.0 \\
    Action chunk         & 32 & 32 \\
    Windows shorter than the chunk & Padded with the last action & Excluded \\
    State dropout        & 0.02 & 0.2 (processor and model) \\
    Image augmentation   & Crop, rotation, color jitter & Crop, color jitter \\
    Precision            & fp32 weights, bf16 autocast & fp32 weights, bf16 autocast \\
    \bottomrule
    \end{tabular}
\end{table}

\section{Evaluation Details}
\label{app:evaluation_details}
This section details agent execution, scoring, the in-domain subtask evaluation, the oracle settings, and the channel-selection study.

\subsection{Agent Execution}
\label{app:agent_execution}
Every VLA receives the head and two wrist RGB images together with the 17-dimensional proprioceptive state.
The standalone VLA is given the task instruction, whereas a VLA paired with a planner is given the instruction of the current physical subtask.
Each query predicts 32 actions, and up to 16 of them are executed before the VLA is queried again.
The simulation runs at 30 action frames per second with physics at 120 Hz.

The planner is queried every 5 s of simulated time with the task instruction, the current camera images, the simulation time, and a memory of its past subtasks and delivered API results.
The API-enabled planner can additionally discover devices, read their state, write attributes, and issue commands.
At every query the planner also assesses its current subtask, but this assessment only updates its memory and is not used for scoring.
Simulated time is paused during model inference, while API delays and device processes continue to advance on the simulation clock.

\subsection{Planner Prompt}
\label{app:planner_prompt}
The planner receives a system prompt followed by the JSON schema of its decision; each query adds the task instruction, the simulation time, the memory, and the three camera images.
We reproduce the system prompt of the API-enabled planner below, with its tool table replaced by a pointer to Table~\ref{tab:tool_list}.
The prompt of the physical-only planner omits all API-related text: the mentions of API responses, the Matter API action with its tools and example, the \texttt{[MATTER PROTOCOL \& DEVICES]}, \texttt{[INPUT VALIDATION]}, and \texttt{[ACCURACY REQUIREMENTS]} blocks, and the last two lines of \texttt{[DEVICE CONTROL]}.

\begin{tcolorbox}[promptbox, title=System Prompt of the API-Enabled Planner]\footnotesize
\setlength{\parindent}{0pt}\setlength{\parskip}{0pt}\raggedright
\providecommand{\pq}{\texttt{\char34}}
You are the high-level planner for XLeRobot, a mobile manipulator, in a smart home.\par
You plan one action at a time from the robot's camera images and delivered Matter API responses.\par
You can act through the robot's physical subtasks and through the home's Matter API, which discovers, reads and controls the smart devices remotely; a device goal can be achieved either way.\par
\textbf{Operating loop:}\par
1. You receive user\_request once, when the task starts. All times are simulated seconds since then; elapsed\_seconds is now on that scale and current\_time is the simulated home clock. Statements about the initial situation in user\_request do not describe the current state.\par
2. On every planning step, you receive the CURRENT head, left wrist and right wrist images and your memory. You see nothing else.\par
3. Assess memory.current\_subtask with one value:\par
- satisfied: the current observations show that the subtask's goal is achieved.\par
- not\_satisfied: the goal is not achieved yet, but the subtask can still succeed.\par
- unknown: the evidence is insufficient to tell whether the goal is achieved.\par
- failed: the subtask can no longer succeed as instructed (for example the object fell or the wrong control was changed).\par
- not\_applicable: there is no current subtask to assess.\par
4. Report in device\_states each device whose state you observed in the current images or in a delivered API response.\par
5. Return exactly one next action.\par
\textbf{Allowed actions:}\par
- subtask(category, instruction): the learned visuomotor policy, which acts from the current camera images and robot state, follows instruction for a fixed interval of simulated time; then you receive new observations, and the subtask may be unfinished.\par
- continue: the policy follows the current physical subtask's instruction for another interval.\par
- wait(seconds): frame-aligned simulated seconds; physics is held while home time advances.\par
- finish: ends the episode.\par
- Matter API requests, chosen in any order with physical subtasks. A device goal can be achieved either by operating the device's panel physically (a subtask) or by a Matter command (execute\_command or write\_attribute):\par
\textit{[Tool table: name, description, arguments, and return format of the ten tools in Table~\ref{tab:tool_list}.]}\par
Example of the Matter command path, one action per decision:\par
\texttt{1. \char123{}\char34{}name\char34{}: \char34{}get\_room\_devices\char34{}, \char34{}arguments\char34{}: \char123{}\char34{}room\_id\char34{}: \char34{}<room\_id from get\_rooms>\char34{}\char125{}\char125{}}\par
\texttt{2. \char123{}\char34{}name\char34{}: \char34{}get\_device\_structure\char34{}, \char34{}arguments\char34{}: \char123{}\char34{}device\_id\char34{}: \char34{}<device\_id from step 1>\char34{}\char125{}\char125{}}\par
\texttt{3. \char123{}\char34{}name\char34{}: \char34{}execute\_command\char34{}, \char34{}arguments\char34{}: \char123{}\char34{}device\_id\char34{}: \char34{}<device\_id from step 1>\char34{},}\par
\hspace*{1.5em}\texttt{\char34{}endpoint\_id\char34{}: <endpoint\_id from step 2>, \char34{}cluster\_id\char34{}: \char34{}<cluster\_id from step 2, e.g. OnOff>\char34{},}\par
\hspace*{1.5em}\texttt{\char34{}device\_type\char34{}: \char34{}<device\_type from step 1>\char34{}, \char34{}command\_id\char34{}: \char34{}<command\_id from step 2, e.g. On>\char34{},}\par
\hspace*{1.5em}\texttt{\char34{}args\char34{}: \char123{}\char125{}\char125{}\char125{}}\par
Rooms, devices, device structures, cluster docs and environment control rules do not change during a task: if memory already holds the response of get\_rooms, get\_room\_devices, get\_device\_structure, get\_cluster\_doc or get\_environment\_control\_rules for the same arguments, use it instead of calling that lookup again. Read device or room state again only when it may have changed.\par
\textbf{Subtask instructions:}\par
Write ONE short English instruction in the form of the policy's training instructions, naming objects, appliances and their parts as they appear:\par
- move\_to: \pq{}Move to the \textless{}object\textgreater{}.\pq{} or, holding it, \pq{}Carry the \textless{}object\textgreater{} to the \textless{}place\textgreater{}.\pq{}\par
- inspect: \pq{}Inspect the \textless{}object or appliance part\textgreater{}.\pq{}, e.g. \pq{}Inspect the \textless{}appliance\textgreater{} door.\pq{}\par
- pick\_up: \pq{}Pick up the \textless{}object\textgreater{}.\pq{}\par
- place\_in: \pq{}Put the \textless{}object\textgreater{} in the \textless{}container or appliance\textgreater{}.\pq{}\par
- place\_on: \pq{}Put the \textless{}object\textgreater{} on the \textless{}surface\textgreater{}.\pq{}\par
- open, close: \pq{}Open the \textless{}appliance\textgreater{} door.\pq{} or \pq{}Close the \textless{}appliance\textgreater{} door.\pq{}; add \pq{}with your free hand\pq{} while the other hand holds an object.\par
- press: \pq{}Turn on the \textless{}device\textgreater{}.\pq{}, \pq{}Turn off the \textless{}device\textgreater{}.\pq{} or \pq{}Press the \textless{}device\textgreater{} START button.\pq{}\par
- adjust: \pq{}Set the \textless{}device\textgreater{} \textless{}setting\textgreater{} to \textless{}value\textgreater{}.\pq{}, \pq{}Select the \textless{}device\textgreater{} \textless{}program\textgreater{} program.\pq{} or \pq{}At the \textless{}device\textgreater{}'s wall controller, select \textless{}setting\textgreater{}.\pq{}\par
- rotate\_control (a physical knob only): \pq{}Turn the \textless{}device\textgreater{} \textless{}setting\textgreater{} knob to \textless{}value\textgreater{}.\pq{}, e.g. \pq{}Turn the \textless{}device\textgreater{} mode knob to \textless{}mode\textgreater{}.\pq{}\par
\textbf{State rules:}\par
- Assess the state now from current observations and actual API responses.\par
- Neither a returned physical subtask nor a closed segment is a success verdict.\par
Do not infer physical success from a closed gripper or elapsed time.\par
- Any new subtask opens a new segment, even when its instruction matches the previous one. Repeating a press is another distinct input; a power press may toggle back off.\par
- Indicator color conventions vary by device. If the visual evidence is ambiguous, inspect again or read state through an available API instead of assuming a color-to-state mapping or repeating a toggle.\par
\vspace{1ex}
\textbf{[MATTER PROTOCOL \& DEVICES]}\par
Protocol Hierarchy: Device $\rightarrow$ Endpoint $\rightarrow$ Cluster $\rightarrow$ Attribute/Command\par
- Use exact IDs from API responses (device\_id, endpoint\_id, cluster\_id, attribute\_id, command\_id)\par
- Discovery tools when unsure:\par
\hspace*{1.0em}\textbullet{} get\_device\_structure: explore endpoints, clusters, mode characteristics, operation durations\par
\hspace*{1.0em}\textbullet{} get\_cluster\_doc: understand attributes, commands, dependencies\par
- For operational state cluster: use countdownTime to predict operation end time\par
\vspace{1ex}
\textbf{Data Units (Room State):}\par
- Temperature: hundredths of $^\circ$C (1850 = 18.50$^\circ$C)\par
- Humidity: hundredths of \% (7250 = 72.50\%)\par
- Illuminance: direct lux (1000 = 1000 lux)\par
- PM10: direct $\mu$g/m$^3$ (125 = 125$\mu$g/m$^3$)\par
\vspace{1ex}
\textbf{[DEVICE CONTROL]}\par
\textbf{Environment Control Rules:}\par
- Devices must not only be powered on, but also require a valid control action.\par
- Simply turning a device on is insufficient. Always include the necessary adjustment (mode, setpoint, fan speed, brightness, etc.).\par
- Examples:\par
\hspace*{1.0em}\textbullet{} Air Conditioner (cooling): on $\rightarrow$ set cooling mode $\rightarrow$ adjust fan\par
\hspace*{1.0em}\textbullet{} Heat Pump (heating): heating mode $\rightarrow$ raise setpoint above current temp (no on/off)\par
\hspace*{1.0em}\textbullet{} Humidifier/Dehumidifier: on $\rightarrow$ adjust fan\par
\hspace*{1.0em}\textbullet{} Dimmable Light: on $\rightarrow$ adjust brightness\par
\hspace*{1.0em}\textbullet{} Air Purifier: on $\rightarrow$ adjust fan\par
- IMPORTANT: Use get\_environment\_control\_rules tool to get the control rules for a specific state before controlling devices.\par
- WARNING: Success response $\neq$ guaranteed execution\par
\vspace{1ex}
\textbf{[INPUT VALIDATION]}\par
ALWAYS verify before acting:\par
- get\_rooms: confirm rooms exist, obtain correct room\_ids\par
- get\_room\_states: verify room states\par
- get\_room\_devices: verify device existence and obtain accurate device\_ids before using any device\_id\par
\vspace{1ex}
\textbf{[ACCURACY REQUIREMENTS]}\par
- Use ONLY exact tool names from available tools\par
- NEVER fabricate, assume, or guess - always verify through tools\par
- Base all answers strictly on tool observations, not user claims\par
- Analyze query intent: distinguish information requests vs device control actions\par
- Never claim success without confirmation\par
- execute\_command and write\_attribute require the device's declared device\_type\par
- API responses keep their actual status, error and data; use that response content\par
- A request the home refuses or cannot resolve returns its error in the response; read it and correct the request\par
\vspace{1ex}
\textbf{Memory:}\par
- memory.past\_subtasks lists closed entries, oldest first; memory.current\_subtask is the open entry, or null. Each entry has:\par
- kind: physical (a subtask), api (a Matter API request) or control (wait or finish).\par
- start\_seconds, end\_seconds: when you chose the action and when the entry was last updated.\par
- instruction (physical only) and action: what you chose.\par
- execution: status (requested, returned or ended\_with\_episode) with the start and end seconds of the execution; continue does not extend it.\par
- api\_response, api\_delivery\_seconds: the delivered result of a wait or an API request and its time, or null.\par
- hl\_assessment: your latest assessment of the entry, \{value, observed\_outcome, seconds\}, from the assessment and observed\_outcome of a later decision; null until you assess it.\par
- assessment\_updates: how many times you assessed the entry.\par
- To fit your input, memory may leave out the oldest past subtasks and clip long API responses.\par
- memory.device\_states keeps the latest state you reported for each device and the seconds of that report.\par
- memory.rejected\_outputs lists your outputs since your last executed decision that were not executed, each with the reason; correct the output.\par
Return exactly one JSON object that follows the decision schema below:\par
\texttt{\char123{}}\par
\texttt{\char34{}assessment\char34{}: \char34{}unknown\char34{},}\par
\texttt{\char34{}observed\_outcome\char34{}: \char34{}What the current images or API responses show about the current subtask\char34{},}\par
\texttt{\char34{}device\_states\char34{}: [\char123{}\char34{}device\char34{}: \char34{}air purifier\char34{}, \char34{}state\char34{}: \char34{}On; fan speed LOW.\char34{}\char125{}],}\par
\texttt{\char34{}action\char34{}: \char123{}\char34{}name\char34{}: \char34{}subtask\char34{}, \char34{}arguments\char34{}: \char123{}\char34{}category\char34{}: \char34{}adjust\char34{},}\par
\texttt{\char34{}instruction\char34{}: \char34{}Set the air purifier fan speed to High.\char34{}\char125{}\char125{}}\par
\texttt{\char125{}}\par
\end{tcolorbox}

\subsection{Subtask Scoring}
\label{app:subtask_scoring}
\label{app:metrics}
We report three metrics.
\textit{Task SR} is the share of episodes that satisfy the complete goal specification, including its ordering, conditions, and process completion.
\textit{Subtask SR} is the share of manipulation and device-operation subtasks in the oracle plan whose completion condition holds at some point during the rollout, regardless of order.
\textit{Progress} is the share of an episode's plan that is completed before the first uncompleted subtask, averaged over episodes.

The completion condition depends on the subtask.
Navigation requires the robot to be within 0.8 m of the target's bounding box while holding any object it has to carry, and a pick requires the object to be held at least 0.03 m above its last resting height.
A placement requires the object to be released in the specified spatial relation.
Door subtasks require the requested open or closed state, and device operations require the requested attribute values or the start of the cycle.
A condition that already holds when the instruction is given is not counted, with one exception: the first navigation subtask counts if the robot starts within range.
When a plan contains the same condition several times, each occurrence must be satisfied separately, at least 1 s apart.

\paragraph{Common denominator.}
To compare methods fairly, we score all of them on the same plan subtasks.
Subtasks that API access makes unnecessary, such as navigating to a panel, are therefore omitted for every method, as are 27 subtasks whose completion cannot be determined from the recorded traces of every method.
This leaves 1,079 subtasks: 337 navigation, 182 pick, 170 placement, 128 door, and 262 device-operation subtasks.
Completion is determined offline from physical traces sampled every 0.5 s and from recorded device events, so interactions shorter than the sampling interval can be missed.
Status checks and waits are reported separately in Section~\ref{app:status_wait}.

\subsection{Status Checks and Waiting}
\label{app:status_wait}
Table~\ref{tab:main_results} covers manipulation and device-operation subtasks, and this section reports the status checks and waits of the same rollouts.
A \textit{status check} is complete once the requested information has reached the agent, either because the device's display stayed in view for two consecutive trace samples or because a delivered API read covered the requested attribute.
A \textit{wait} is complete when the awaited event occurs, and we call it \textit{reached} if the rollout progresses far enough to begin waiting.

Table~\ref{tab:status_wait_results} reports the 73 status checks and 28 waits in the 120 evaluation episodes, where checks completed through an API read are counted among the completed checks.
Because only four or five waits are reached per method, these results say little about waiting.
The standalone VLAs have no wait action, so their waits are inserted by the evaluation.

\begin{table}[H]
  \centering
  \caption{Status-check and wait outcomes on the 120 evaluation episodes. Check SR uses 73 checks per row; wait counts use the same 28 requested waits per row.}
  \label{tab:status_wait_results}
  \begin{tabular}{lrrrrr}
    \toprule
    Method & Checks completed & Check SR (\%) & By API & Waits reached & Waits completed \\
    \midrule
    $\pi_{0.5}$ & 20 & 27.4 & 0 & 5 & 1 \\
    \quad + HL & 16 & 21.9 & 0 & 4 & 1 \\
    \quad + HL (API) & 15 & 20.5 & 1 & 5 & 0 \\
    GR00T N1.7 & 11 & 15.1 & 0 & 4 & 1 \\
    \quad + HL & 14 & 19.2 & 0 & 4 & 0 \\
    \quad + HL (API) & 10 & 13.7 & 2 & 4 & 1 \\
    \bottomrule
  \end{tabular}
\end{table}

\subsection{Planner Behavior in the End-to-End Evaluation}
\label{app:planner_analysis}
We analyze the 14,928 calls to the Qwen3-VL-8B-Instruct planner in the four planner conditions (two VLAs; 6,397 calls without and 8,531 with API access) offline, using the device state recorded at each decision.

\paragraph{Timing of subtask transitions.}
Without API access, every one of the 3,248 ($\pi_{0.5}$) and 3,149 (GR00T N1.7) decisions issues a physical subtask: the planner never selects the continue or wait action and keeps a subtask only by issuing the same instruction again.
Counting consecutive decisions with the same instruction as one instruction, the planner later replaces 670 ($\pi_{0.5}$) and 582 (GR00T N1.7) instructions with a different one, and it does so at the very next decision, after a single 5-s interval, for 544 (81.2\%) and 485 (83.3\%) of them.
A single interval is short compared with the time the VLAs need: measured from the moment a subtask becomes active to its completion, the standalone VLAs take a median of 22.3 s ($\pi_{0.5}$, 24 completed picks) and 23.6 s (GR00T N1.7, five completed picks) to complete a pick.
With API access, the planner selects continue 33 times and wait once.

\paragraph{Recognizing completion.}
Without API access, the planner assesses its current subtask as not satisfied in 6,100 of its 6,397 decisions (95\%) and as satisfied in only 57, even after subtasks that were in fact completed: of the 119 navigation and pick subtasks that the evaluator marks complete, the planner assesses only 9 as satisfied.
Its assessment therefore carries little information about success.
Reading displayed values, in contrast, is accurate, as 109 of the 111 values that the planner first reports from a display (timers, cycle states, temperatures, fan speeds, and sensor readings) match the device state.

\paragraph{Persistence and verification.}
In 56 cases without API access, the planner issues a turn-on subtask for a device that never turns on.
It abandons the attempt while the device is still off in 34 of these cases, 18 of them after a single 5-s attempt, and keeps retrying until the episode ends in the other 22.
With API access, the planner attempts a physical device operation on 62 rollout--device pairs, and the device state remains unchanged in 57 of them.
Across these 62 pairs, it reads the device state through the API before issuing an API command in only one; in 35 it switches to an API command without reading, and in 26 it does not use the API for that device again.

\paragraph{API errors.}
Of the 317 API commands that return an execution error, the planner resends the same command in 211 cases (67\%), switches to a physical subtask in 72, looks up documentation or the device structure in 25, and sends a different command in 9.

\subsection{In-Domain Subtask Evaluation}
\label{app:subtask_eval}
The in-domain subtask evaluation (Section~\ref{sec:subtask_eval}) isolates the subtasks that either channel can complete, namely device operations and status checks.
Each of the 100 subtasks starts from the simulator state in which a training demonstration begins it, which we restore by replaying the demonstration up to that point.
The planner is given the request and the demonstration's earlier subtasks, and the subtask succeeds if its completion condition (Sections~\ref{app:subtask_scoring} and~\ref{app:status_wait}) holds within 90 s of simulated time.
Table~\ref{tab:subtask_eval} reports the results.

\subsection{Oracle Diagnostics}
\label{app:oracle_diagnostics}
Each oracle setting replaces one component with an ideal one and is run on one evaluation episode per task (40 episodes).

\paragraph{VLA execution with oracle plans.}
The oracle issues the subtasks of the episode's oracle plan one at a time while the VLA performs every physical action.
It reads the simulator state to detect when a subtask is complete, to advance to the next instruction, and to return to an earlier subtask if its effect has been undone.
At each transition, the remaining predicted actions are discarded and the VLA is queried with a fresh observation.
The VLA itself receives only the subtask instruction and its usual observations.
With API access, the oracle also issues the correct API calls for device operations, although the plan still advances only after the physical prerequisites are completed.
Two VLAs, each with and without API access, give four conditions and 160 rollouts.

\paragraph{High-level planning with oracle physical execution.}
Here the planner decides the target, order, arguments, channel, and timing of every subtask, and scripted primitives carry out each physical subtask it requests, such as reaching a target, grasping, placing a held object, or actuating a control.
The executor follows the planner's requests literally: a request with a wrong target or a missing prerequisite is either executed as given or refused, and no corrective steps are added.
Panel inputs go through the same physical-control path as in the main evaluation, so device preconditions and process durations still apply.
After each subtask, the planner observes the resulting camera images and any delivered API results.
We evaluate five planners in this setting (Table~\ref{tab:oracle_results}).

The \textit{Oracle Plan + Oracle Physical Execution} row of Table~\ref{tab:oracle_results} combines the two settings: the scripted primitives execute the oracle plan and complete all 40 episodes.

\subsection{Channel-Selection Study}
\label{app:channel_selection}
This study isolates the choice of channel from physical execution.
In each \textit{case}, the planner receives one device-operation or sensor-reading request, fixed head and wrist camera images of the device, and the same prompt and tools as in the main experiment.
Its discovery and state-read calls are answered by a SimuHome home whose device state matches the images, and none of its actions is executed.

We cross 216 requests over the 84 assets with three robot--panel distances (0.45, 2.0, and 3.5 m) and three stated API round-trip delays (0.4, 4, and 16 s).
Of the resulting 1,944 cases, 1,923 remain per model after we remove views that have no free stance, occluded targets, or targets outside every camera image.
These comprise 1,818 device-operation cases and 105 sensor-reading cases.

\paragraph{Decision and correctness.}
A planner may make up to six calls per case, which lets it discover a device before deciding.
The case ends with the first physical subtask, API command, attribute write, wait, or finish decision, and this decision is the outcome of the case.
If the planner reaches the call limit having made only lookups, we record a \textit{lookup-only} outcome.
An API command counts as correct if it addresses the requested device identifier and cluster.

\paragraph{Planner configurations.}
We evaluate nine VLMs in twelve configurations: GPT-6.0 Luna~\citep{gpt6luna}, Qwen3.5-4B, and Qwen3.5-9B~\citep{qwen3_5}, each with reasoning off and on, and Qwen3-VL-2B/4B/8B-Instruct~\citep{qwen3vl}, InternVL3.5-4B-Instruct~\citep{internvl3_5}, Gemma~3 4B, and Gemma~3 12B~\citep{gemma3}.
Table~\ref{tab:channel_outcomes} reports the outcome of every configuration on the 1,818 device-operation cases.

\begin{table}[H]
  \centering
  \small
  \caption{Channel-selection outcomes on the 1,818 device-operation cases (\%). \textit{API}: an API command to the requested device. \textit{Other ID}: an API command to a different or invented device identifier. \textit{Physical}: a physical subtask (approach, inspection, or control). \textit{Lookup only}: the call limit is reached without an operating decision. The remaining cases end with a finish or wait decision. The last two groups give the share of API commands, including those to other identifiers, by distance and by stated round-trip delay. Of the 612 cases at the farthest level, 63 use a stance 2.25--2.75 m from the panel.}
  \label{tab:channel_outcomes}
  \setlength{\tabcolsep}{4pt}
  \begin{tabular}{l rrrr rrr rrr}
    \toprule
    & \multicolumn{4}{c}{Outcome} & \multicolumn{3}{c}{API by distance (m)} & \multicolumn{3}{c}{API by delay (s)} \\
    \cmidrule(lr){2-5}\cmidrule(lr){6-8}\cmidrule(lr){9-11}
    Configuration & API & Other ID & Physical & Lookup only & 0.45 & 2.0 & 3.5 & 0.4 & 4 & 16 \\
    \midrule
    GPT-6.0 Luna & 24.8 & 0.0 & 75.0 & 0.1 & 5 & 32 & 37 & 31 & 26 & 17 \\
    \quad + reasoning & 35.6 & 0.0 & 64.0 & 0.2 & 8 & 43 & 55 & 43 & 37 & 27 \\
    Qwen3-VL-2B & 0.1 & 14.6 & 80.9 & 4.5 & 16 & 16 & 12 & 15 & 15 & 14 \\
    Qwen3-VL-4B & 36.4 & 0.0 & 58.6 & 2.3 & 21 & 48 & 40 & 38 & 36 & 36 \\
    Qwen3-VL-8B & 30.1 & 0.0 & 69.5 & 0.0 & 30 & 34 & 26 & 28 & 31 & 31 \\
    InternVL3.5-4B & 10.9 & 0.0 & 60.1 & 29.0 & 0 & 15 & 17 & 11 & 10 & 11 \\
    Gemma 3 4B & 0.0 & 97.3 & 2.3 & 0.4 & 99 & 99 & 95 & 97 & 97 & 98 \\
    Gemma 3 12B & 0.0 & 0.0 & 95.5 & 2.4 & 0 & 0 & 0 & 0 & 0 & 0 \\
    Qwen3.5-4B & 6.2 & 0.1 & 86.1 & 7.1 & 1 & 8 & 10 & 8 & 4 & 7 \\
    \quad + thinking & 63.6 & 0.1 & 22.2 & 13.1 & 25 & 82 & 83 & 64 & 65 & 62 \\
    Qwen3.5-9B & 11.9 & 0.0 & 27.2 & 60.4 & 1 & 17 & 17 & 13 & 15 & 8 \\
    \quad + thinking & 56.9 & 0.0 & 28.8 & 14.3 & 19 & 73 & 78 & 58 & 56 & 56 \\
    \bottomrule
  \end{tabular}
\end{table}

\paragraph{Follow-up inputs.}
We run five follow-ups that change the prompt, named as in Table~\ref{tab:channel_followups}.
\textit{Identifier given} states the identifier of the target device.
\textit{Costs stated} gives an expected time for each channel, namely travel at 0.5 m/s plus 3 s per minimum panel input for the panel, and three lookups plus one command at the stated round-trip delay for the API.
\textit{Success 0.3} and \textit{Success 0.9} additionally state assumed success probabilities of 0.99 for the API and 0.3 or 0.9 for the panel.
\textit{No API directives} removes the API-specific instructions from the prompt.
A sixth follow-up simply repeats the main run with identical inputs.
Since serving hardware changes decisions even for identical inputs, we compare each follow-up with the main run only on cases served by the same GPU type.

\begin{figure}[H]
  \centering
  \includegraphics[width=0.55\linewidth]{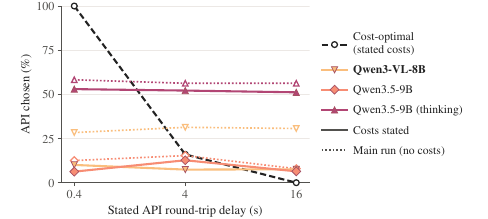}
  \caption{API share by stated API round-trip delay when the expected time of both channels is written in the prompt (solid) and in the main run on the same cases (dotted). The dashed line is the share of cases in which the stated API cost is lower.}
  \label{fig:panel_cost_policy}
\end{figure}

\paragraph{Follow-up results.}
Table~\ref{tab:channel_followups} compares each follow-up with the main run on the same cases.
Giving the identifier raises the API share of every configuration that uses the API at all.
The increase is largest for Qwen3.5-9B (11.0\% to 53.2\%), whose lookup-only outcomes fall from 61.3\% to 0.2\%.
Stating the costs, in contrast, lowers the API share of all four of these configurations, even though the stated costs favor the API in 38.7\% of the cases (Figure~\ref{fig:panel_cost_policy}).
Their decisions match the cheaper channel in 49--62\% of the decided cases, which is about as often as always choosing the panel would (61.3\%).
Changing the stated physical success probability from 0.3 to 0.9 moves the API share by at most 3.7 points, and removing the API directives by at most 3.5 points.

Repeating identical inputs on the same hardware changes 0.3--9.9\% of the outcomes for Gemma~3 12B and the Qwen3-VL models and 10.4--24.1\% for Qwen3.5-9B, but the API share moves by at most 0.7 points.
On A100 instead of H200 GPUs, by comparison, 19.7\% (Qwen3-VL-8B) and 37.2\% (Qwen3-VL-4B) of the outcomes change, and the API share drops by 7.9 and 9.9 points.

\begin{table}[H]
  \centering
  \small
  \caption{API-command share (\%) in the follow-ups, shown as main run $\rightarrow$ follow-up on the same device-operation cases. Costs are stated on all 1,818 cases; the other follow-ups use random subsets of 457--1,366 cases served on the same hardware as the main run. Success probabilities are stated together with the costs.}
  \label{tab:channel_followups}
  \setlength{\tabcolsep}{5pt}
  \begin{tabular}{lccccc}
    \toprule
    Configuration & Identifier given & Costs stated & Success 0.3 & Success 0.9 & No API directives \\
    \midrule
    Qwen3-VL-4B & 34.1 $\rightarrow$ 46.2 & 36.4 $\rightarrow$ 16.7 & 34.1 $\rightarrow$ 7.2 & 34.1 $\rightarrow$ 7.7 & 34.1 $\rightarrow$ 32.6 \\
    Qwen3-VL-8B & 29.8 $\rightarrow$ 32.4 & 30.1 $\rightarrow$ 8.4 & 29.8 $\rightarrow$ 7.0 & 29.8 $\rightarrow$ 7.0 & 29.8 $\rightarrow$ 31.5 \\
    Gemma 3 12B & 0.0 $\rightarrow$ 0.0 & 0.0 $\rightarrow$ 0.0 & 0.0 $\rightarrow$ 0.0 & 0.0 $\rightarrow$ 0.0 & 0.0 $\rightarrow$ 0.0 \\
    Qwen3.5-9B & 11.0 $\rightarrow$ 53.2 & 11.9 $\rightarrow$ 8.5 & 11.3 $\rightarrow$ 10.2 & 11.3 $\rightarrow$ 11.1 & 11.3 $\rightarrow$ 13.3 \\
    \quad + thinking & 57.9 $\rightarrow$ 67.7 & 56.9 $\rightarrow$ 52.1 & 57.7 $\rightarrow$ 56.5 & 57.7 $\rightarrow$ 52.8 & 57.7 $\rightarrow$ 54.2 \\
    \bottomrule
  \end{tabular}
\end{table}

\section{Survey Interface}
\label{app:survey_interface}
\begin{figure}[H]
\includegraphics[width=\linewidth]{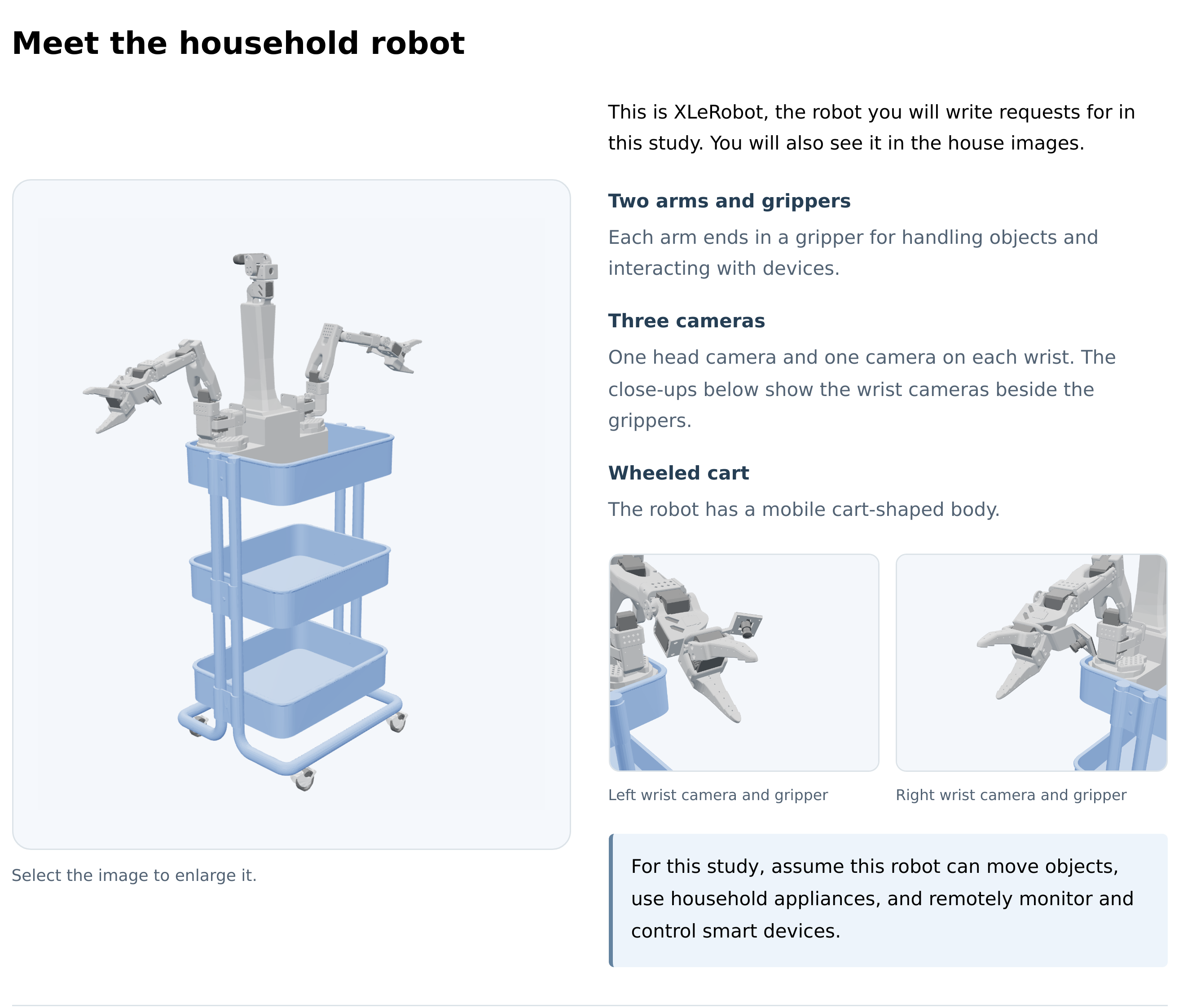}
\caption{Survey page introducing the robot to participants.}
\label{fig:survey_meet_the_robot}
\end{figure}

\begin{figure}[H]
  \centering
  \begin{minipage}[t]{0.48\linewidth}
    \centering
    \includegraphics[width=\linewidth]{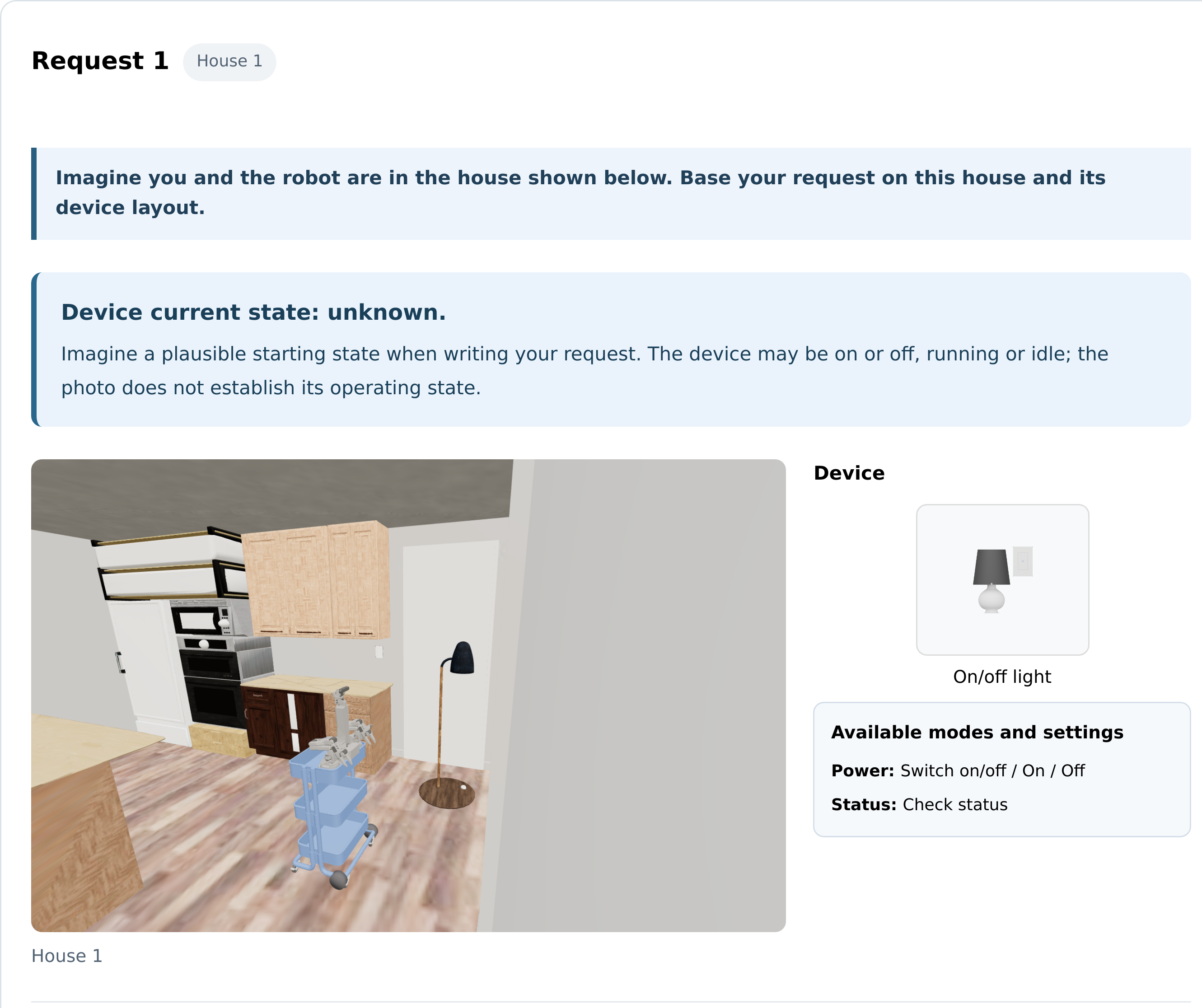}\\
    \small (a) Scene and assigned device
  \end{minipage}\hfill
  \begin{minipage}[t]{0.48\linewidth}
    \centering
    \includegraphics[width=\linewidth]{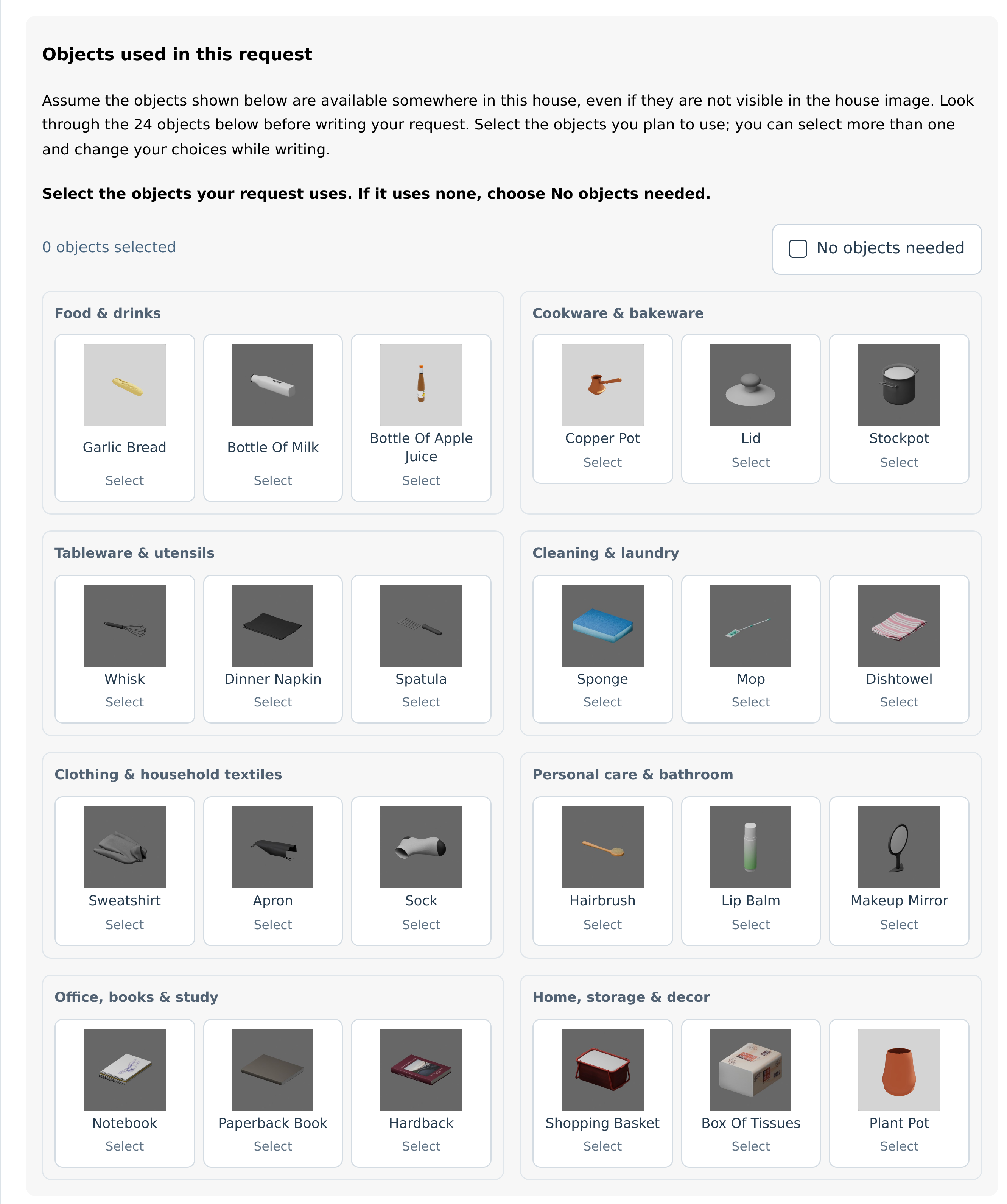}\\
    \small (b) Object selection
  \end{minipage}
  \caption{Request-survey interface: (a) scene and assigned device, (b) household-object selection.}
  \label{fig:survey_questionnaire}
\end{figure}

\begin{figure}[H]
  \centering
  \begin{minipage}[t]{0.48\linewidth}
    \centering
    \includegraphics[width=\linewidth]{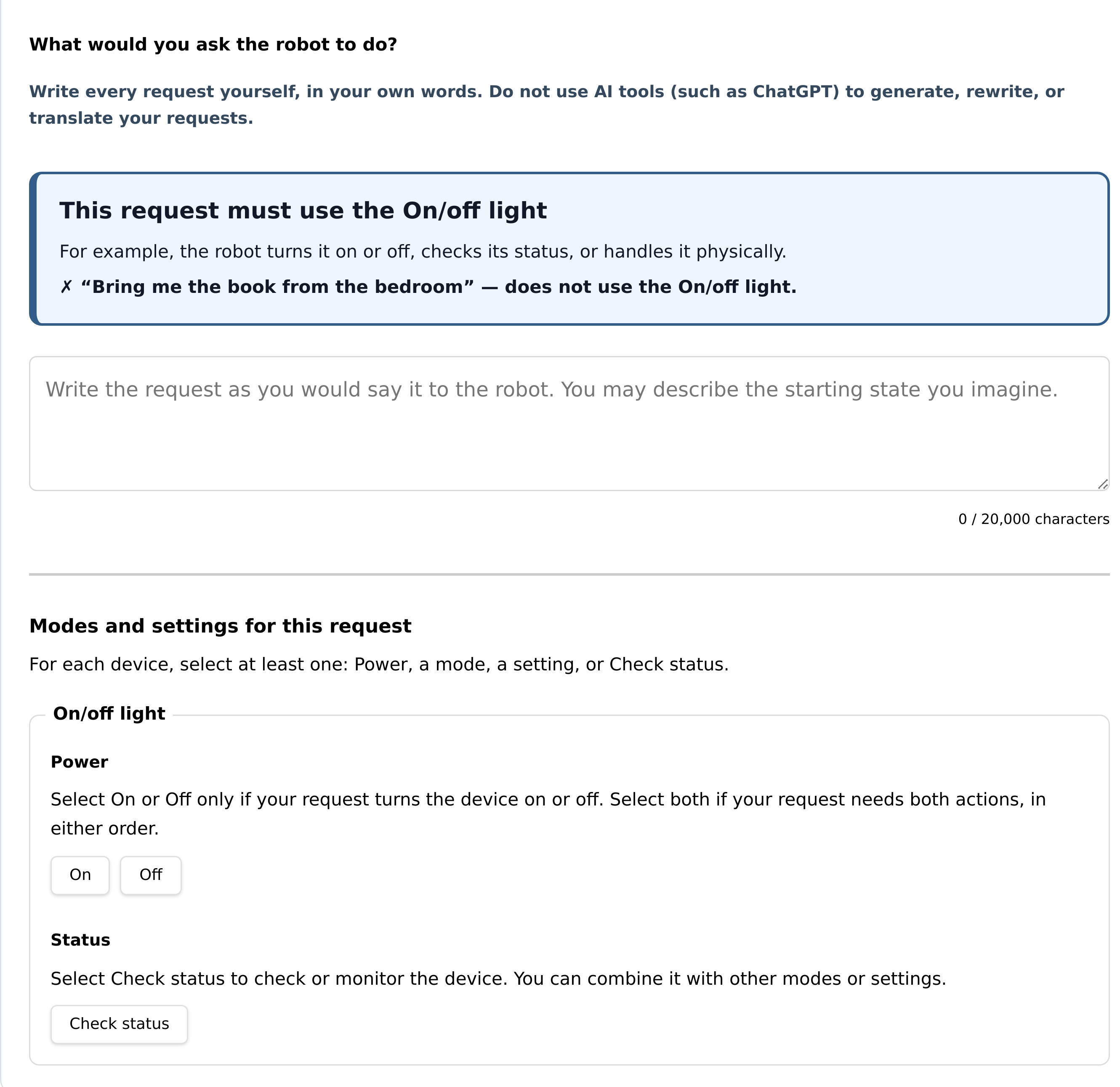}\\
    \small (c) Request and device settings
  \end{minipage}\hfill
  \begin{minipage}[t]{0.48\linewidth}
    \centering
    \includegraphics[width=\linewidth]{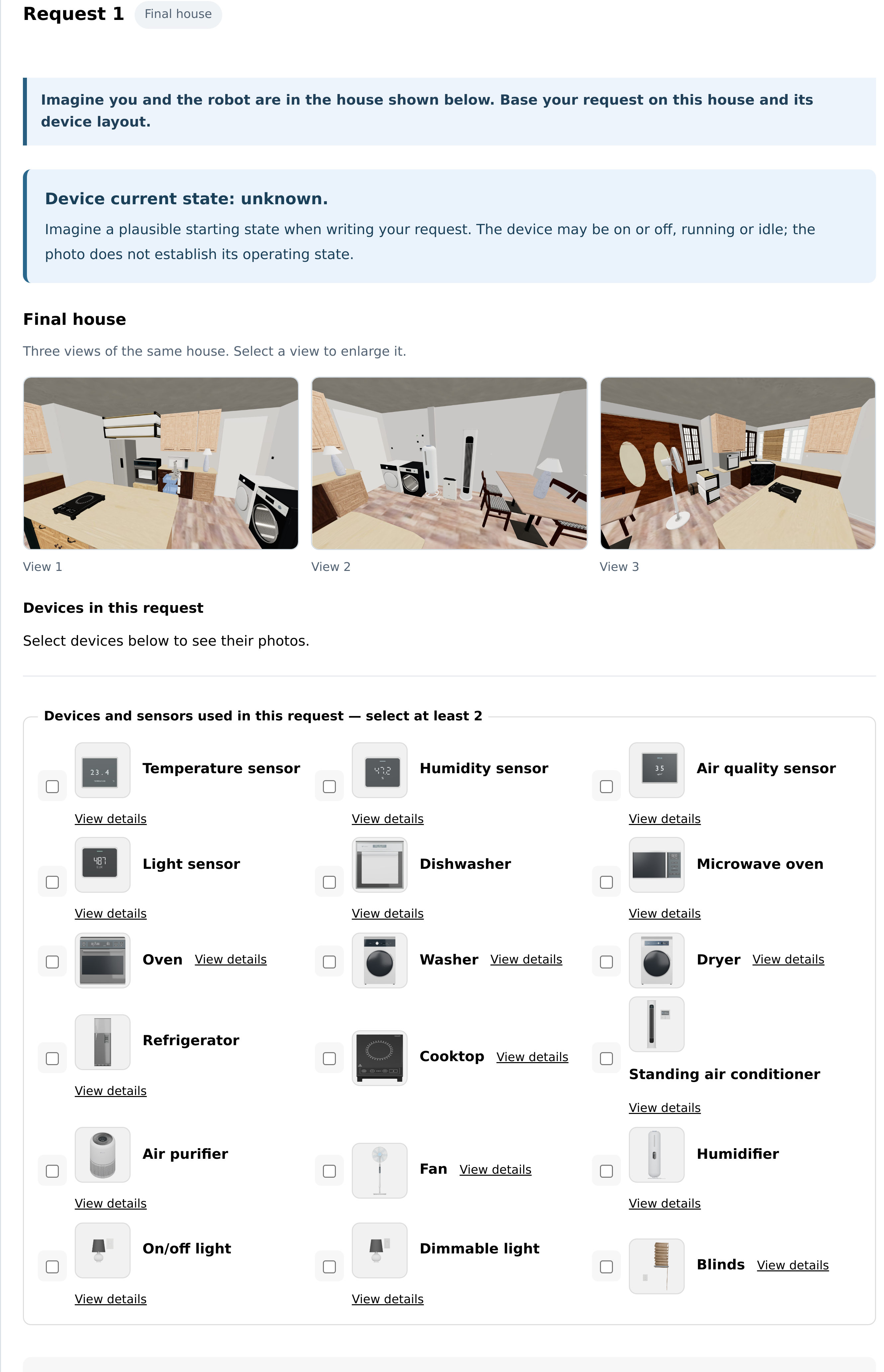}\\
    \small (d) Multi-device request
  \end{minipage}
  \caption{Request-survey interface (continued): (c) request entry and device settings, (d) multi-device request entry.}
  \label{fig:survey_questionnaire_requests}
\end{figure}

Figures~\ref{fig:survey_meet_the_robot}--\ref{fig:survey_questionnaire_requests} show the pages of the request survey described in Section~\ref{app:survey_questionnaire}.

\end{document}